\documentclass[preprint,12pt]{elsarticle}

\usepackage{amsmath,amssymb,amsfonts, amstext}
\usepackage{longtable,booktabs}
\usepackage{array}
\usepackage{lscape} 
\usepackage{float}
\usepackage{multirow} 
\usepackage{nicefrac} 
\usepackage{subcaption}
\usepackage{accents}
\newcommand{\tabitem}{~~\llap{\textbullet}~~}
\DeclareMathOperator*{\argmin}{argmin}

\journal{Information Sciences}

\begin{document}

\begin{frontmatter}

\title{Using Shape Constraints for Improving Symbolic Regression Models}

\author[hag]{C. Haider\corref{correspondingauthor}}
\cortext[correspondingauthor]{Correspoding author}
\ead{christian.haider@fh-hagenberg.at}

\author[bra]{F. O. de Franca\fnref{fabricio}}
\fntext[fabricio]{Email adress: folivetti@ufabc.edu.br}

\author[hag]{B. Burlacu}

\author[hag]{G. Kronberger}

\affiliation[hag]{organization={Josef Ressel Center for Symbolic Regression, University of Applied Sciences Upper Austria},
            addressline={Softwarepark 11},
            city={Hagenberg},
            postcode={4232},
            country={Austria}}

\affiliation[bra]{organization={Center for Mathematics, Computation and Cognition (CMCC), Heuristics, Analysis and Learning Laboratory (HAL),
Federal University of ABC},
            city={Santo Andre},
            country={Brazil}}



\begin{abstract}
    We describe and analyze algorithms for shape-constrained symbolic
    regression, which allows the inclusion of prior knowledge about
    the shape of the regression function. This is relevant in many areas of
    engineering -- in particular whenever a data-driven model obtained from
    measurements must have certain properties (e.g. positivity,
    monotonicity or convexity/concavity).
    We implement shape constraints using a soft-penalty approach which uses
    multi-objective algorithms to minimize constraint violations and training
    error. We use the non-dominated sorting genetic algorithm (NSGA-II) as well
    as
    the multi-objective evolutionary algorithm based on decomposition (MOEA/D).
    We use a set of models from physics textbooks to test the algorithms and compare
    against earlier results with single-objective algorithms. The results show
    that all algorithms are able to find models which conform to
    all shape constraints.
    Using shape constraints helps to improve extrapolation
    behavior of the models.
\end{abstract}




\end{frontmatter}


\section{Introduction}
\label{sec:introduction}

Modeling of dynamic and complex systems often requires adherence to certain physical behaviors to ensure that the resulting model is trustworthy. A model that adheres to the desired physical laws is essential when studying the control of physical systems since it ensures that the resulting model has a predictable behavior and can be validated and understood by domain experts. Due to the high complexity of such systems, it is usually impossible to model them based on first principles alone, thus requiring empirical data-based modeling. 
The integration of physical domain knowledge with data-based empirical modeling is a relatively new and not fully-explored, although increasingly important area of research~\cite{baker2019workshop}.

Symbolic regression (SR) searches for models expressed as formulas whereby the
algorithm identifies the model structure and parameters simultaneously. Even though the approach is data-based, it opens the possibility to find short expressions similar to white-box models derived from first principles. Evolutionary algorithms represent a well-known approach for this task.
In~\cite{ourPaper}, SR was further extended to constrain the search space with models that have desired physical properties. We have described such constraints by measuring properties of the shape of the model, such as monotonicity and convexity, using interval arithmetic. We proposed shape-constrained symbolic regression (SCSR)~\cite{ourPaper} as a supervised learning approach for finding SR models.

\subsection{Objectives}
\label{sbusec:problemStatement}
In this paper, we extend~\cite{ourPaper} by using multi-objective approaches to find a set of solutions that minimize both approximation error and soft-penalties incurred by constraint violations. Furthermore, we investigate how the extrapolation behavior is affected by shape constraints. We hypothesize that shape constraints can help find more accurate regression models with improved extrapolation ability, especially under the presence of noise in the training data.

The broader goal of this project is to integrate domain knowledge into SR
algorithms to ensure that the returned model has a better correspondence to the actual
underlying physical conditions of the system.

Specifically for this paper, we have two objectives. The first one is to improve
the search process of a constrained model using multi-objective algorithms.
The second one is to investigate whether a conforming model has additional
benefits of mitigating the effect of noise in the training data and improving extrapolation behavior.

\subsection{Assumptions}
\label{sbusec:assumptions}

We assume that we have potentially noisy measurements from a physical system.
The obtained data may not be representative of the whole domain. Additionally
we assume that knowledge about the system's expected behavior can
be described by shape constraints. A regression model shall be identified using
supervised learning guided by knowledge.

This paper is organized as follows, in
Section~\ref{sec:shapeConstrainedRegression} we describe the concept of shape
constraints and how to check the feasibility of a SR model using interval arithmetic. In
Section~\ref{sec:relatedWork} we summarize related work. We describe our approach and the single-objective
and multi-objective algorithms in Section~\ref{sec:method}.
Sections~\ref{sec:analysisInDomain}~and~\ref{sec:analysisOutOfDomain} describe
the experimental setup, problem instances, and the obtained results. Finally, we
discuss the results in Section~\ref{sec:discussion} and conclude the paper in
Section~\ref{sec:conclusion}.

\section{Shape-constrained Regression}
\label{sec:shapeConstrainedRegression}

Shape constraints allow to enforce desired properties of a regression (or
classification) model. For example, we might know that a function can be
monotonically increasing w.r.t. a specific variable, the target must be
positive, or that the model should be convex. Mathematically, we can express
these properties using partial derivatives of the model.
Table~\ref{tab:shapes} lists the constraints that we consider. Higher-order
derivatives are possible as well but not considered in this work. The idea of shape
constraints is not novel (cf.~\cite{Wright1980}), but
it has recently received increasing attention in the machine learning field, as
we will detail in the next section.

\begin{table}
    \center
    \caption{Shape constraints used in the SR algorithms. All constraints assume a domain $l_i \leq x_i \leq u_i$ for each variable $x_i$.}
    \label{tab:shapes}
    \begin{tabular}{lr}
       \hline
        Property & Mathematical formulation \\
        \hline
        Non-negativity & $f(x) \geq 0$ \\
        Non-positivity & $f(x) \leq 0$ \\
        Image inside a boundary & $l \leq f(x) \leq u$ \\
        Monotonically non-decreasing & $\frac{\partial}{\partial x_i}f(x) \geq 0$ \\
        Monotonically non-increasing & $\frac{\partial}{\partial x_i}f(x) \leq 0$ \\
        Convexity & $\frac{\partial^2}{\partial x_i^2}f(x) \geq 0$ \\
        Concavity & $\frac{\partial^2}{\partial x_i^2}f(x) \leq 0$ \\
        \hline
    \end{tabular}
\end{table}

Although we can describe shape constraints unambiguously, their evaluation often
requires the use of approximation methods. Consider the non-decreasing
monotonicity constraint; it requires that the partial derivative of the model w.r.t. the
variable $x_i$ is equal or greater than zero for the whole input domain. To
check this, we have to find
the minimum value of the partial derivative in the given domain. If the model is
non-linear, this implies a non-linear optimization problem that is often
NP-hard. Approximations that can be solved more efficiently are therefore useful.

Approximations for the evaluation of the shape constraints can be classified as
optimistic or pessimistic~\cite{Gupta2016}. Optimistic approximations check the constraints
for a finite number of points in the input space. If the choice of points is
not representative, they can accept infeasible solutions. To
alleviate this problem, a large number of samples is required. Because of that,
these approximations do not scale well with problem dimensionality.
Pessimistic approximations guarantee the acceptance only of feasible solutions.
This is more conservative and can lead to rejection of feasible
solutions.

\citet{Gupta2016} identified different strategies to cope with monotonicity constraints: 
i) constraining the closed-form model (i.e., only positive coefficients), 
ii) pruning constraint violations after adjusting the model,
iii) penalization of violations during optimization, 
iv) re-labeling the training data to be monotonic. Besides these strategies, some proposals apply constrained optimization where the constraints
depend on a sample of the training data. The sampling and penalization strategies are \emph{optimistic} as they trust that the samples are representative. 
However, this can lead to a false assumption that a model is feasible. 
Strategies i, ii, and iv guarantee that the model respects the constraints, but they may limit the search space, impacting the accuracy.

We apply a pessimistic approach using interval arithmetic (IA) to
calculate bounds for the model and partial derivatives as explained in
Section~\ref{subsec:intervalArihmetic}. IA has a smaller computational cost than the
evaluation of model error and its execution cost does not increase the
asymptotic runtime complexity of the algorithms.

\begin{equation}
    \begin{aligned}
    \label{equation:shapeConstrainedRegression}
        f^{*}(x) = \argmin_{f(x) \in \mathcal{M}} L(f(x), y), \qquad x \in \Omega\\
        \text{subject to shape constraints }~c_{i}(\mathcal{X}_{i}), \qquad \mathcal{X}_{i} \subseteq \Omega.
    \end{aligned}
\end{equation}

Equation \ref{equation:shapeConstrainedRegression} shows the definition of 
shape constrained regression, where $L(f(x), y)$ is the loss function, $f(x)$ the
model, $y$ the target, $\mathcal{M}$ denotes the model space and $\Omega$ is the
domain and $c_{i}$ are shape constraints as shown in Table
\ref{tab:shapes} with $\mathcal{X}_{i}$ are the domains
for constraints.

\subsection{Interval Arithmetic}
\label{subsec:intervalArihmetic}
IA \cite{IAPrinciples} is a mathematical method that
allows associating the input of a system with an interval instead of a scalar
value \cite{IA-algorithms}. It is commonly used for floating-point arithmetic in
scientific programming to deal with rounding and numerical inaccuracies of
fixed-length floating-point representation or to specify intervals for input
values \cite{IntervalAnalysis}.

The idea behind IA is that instead of a scalar value $x$, an interval $[a,
b]$ is defined, such that $[a, b] = \{x \in \mathbb{R} \mid a \leq x \leq b\}$ where $a$ and
$b$ are named lower bound and upper bound. Mathematical operations are defined over intervals. An operation $\langle
op \rangle$ is defined as $[x_{1}, x_{2}]\; \langle op \rangle\;
[y_{1}, y_{2}] = \{x\; \langle op \rangle\; y \mid x \in [x_{1}, x_{2}]\;
\text{and}\; y \in [y_{1}, y_{2}]\}$. Within a sequence of calculations, defined by a mathematical
expression, all operations are replaced by interval operations 
resulting in an interval that gives upper and lower bounds for the expression result. This calculation can return exact bounds or an overestimation, thus
the returned interval is guaranteed to always include the actual bounds.

This overestimation results from the dependency problem which describes the fact that variables
that occur multiple times are not considered as independent~\cite{IAPrinciples, constraintedIntervalArithmetic}. For instance if the given interval $X =
[a, b]$ is subtracted from itself, the resulting interval will be calculated as
$X - X = [a - b, b - a]$, whereas the correct answer should be $X - X = [0, 0]$.

If we define the interval of each input variable as its domain, we can use IA
to evaluate the shape constraints in Table~\ref{tab:shapes}. For example,
to verify if a model is monotonically non-decreasing w.r.t. a $x_i$,
we evaluate the partial derivative of the model by replacing the input values
with the interval representing their domain. This will result in an interval
$[\underaccent{\bar}{y}, \bar{y}]$. The model will satisfy the constraint if
$\underaccent{\bar}{y} \geq 0$.
If this interval is overestimated, it might prevent the assertion of the 
model's feasibility. However, whenever the evaluation asserts that the model
is feasible, it is guaranteed to be so.


\section{Related Work}
\label{sec:relatedWork}

Although monotonic constraints have been studied in statistics literature for some time~\citep{brunk1972statistical},
it has recently become an important part of eXplainable AI~\citep{baker2019workshop, rai2020explainable}
as it can improve the comprehensibility, fairness, and safety
of the generated models by enforcing models to match expected behavior~\citep{liu2020certified}.

Another name for models with monotonic constraints is isotonic models (order-preserving models)~\citep{brunk1972statistical, Wright1980, Tibshirani2011}.
The common approaches for isotonic models are step functions~\citep{chakravarti1989isotonic}, solving
a convex optimization problem~\citep{Tibshirani2011}, and isotonic splines~\citep{Wright1980}.
Isotonic models are most commonly associated with the Pool of Adjacent Violators
algorithm~\citep{chakravarti1989isotonic} that, for a one-dimensional data set, 
sorts data by increasing input values and groups them in such a way
that the median over a group is smaller or equal than the median over the next group, thus
rewriting the target values of the training data to be monotonic.

A possible solution to enforce monotonicity in parametric models is
to constrain the adjustable parameters to generate only feasible models.
\citet{sill1998monotonic} constrained the coefficients connected to a monotonic variable
of a multi-layered neural network to be positive (or negative if monotonically decreasing)
and then calculates the boundaries of the coefficients for the following layers. 
While this approach guarantees the creation of a monotonic model, it is pessimistic.
This strategy is similar to how XGBoost~\citep{chen2016xgboost}, an extension of
gradient tree boosting~\citep{friedman2001greedy}, ensures monotonicity.
At every split, it propagates an upper and lower limit to the coefficients
of the next splits~\citep{bartley2019enhanced}.

In another approach called monotonic hint, \citet{Sill1997}
introduce a penalty factor into the loss function proportional to the 
amount of constraint violations.
They estimate these violations using the training data. 
As a consequence the approach is optimistic.

Other parametric methods use monotonic hints~\citep{Abu-Mostafa1992, Lauer2008} 
to handle shape constraints. The main advantage of this approach is 
that it is simple to include other constraints (e.g., convexity, positivity)
without changing the main algorithm.

\citet{liu2020certified} introduced a mixed-integer linear programming (MILP) 
formulation to verify the monotonicity of piecewise linear neural networks. 
The main idea is to adjust the coefficients of the neural network using a 
regularized loss function, similar to \citep{Sill1997}, and solving the 
MILP problem to ensure that the adjusted NN is monotonic. 
If not, the algorithm increases the regularization factor and retrains the NN.

For polynomial models the theory around
sum-of-squares (SOS) polynomials enables incorporation of shape
constraints into polynomial regression. The main idea is that a
polynomial is non-negative if it can be represented as a sum of
squared polynomials \citep{parrilo2000structured}.
\citet{hall2018optimization} gives a good introduction to the theory
and describes multi-variate shape-constrained polynomial regression
(SCPR) based on semidefinite programming to find sum-of-squares (SOS)
representations. The main advantage of this approach is that it can
be solved efficiently using SDP solvers. On the other hand, the
problem size grows quickly with the problem dimension and the
polynomial degree. \citet{papp2018sumofsquares} describe an
alternative approach to solve the SOS problem without resorting to
SDP, which can reduce the computational cost to fit an SCPR model
especially for low-dimensional problems. However, this approach has
not yet been analyzed fo SCPR.

Cubic smoothing splines are a piecewise regression model
with continuity constraints at knot points to improve the smoothness of the generated function.
\citet{Ramsay1988} introduces the idea of integrated splines (I-Splines) which
work together with constraints of non-negativity for the coefficients to ensure
a monotonic model.
\citet{Papp2014} formulate this problem as a second-order cone programming (SOCP) problem
to avoid the limitations of the previous method. Both methods only support uni-variate problems.

Lattice Regression~\citep{Gupta2016, Milani2016} creates a $M_1 \times M_2 \times \ldots M_D$ 
(usually $D=2$) lattice, projecting the input space. Each dimension of the lattice is composed 
of $M_i$ keypoints stored in a lookup table. The target value for a sample point is estimated
by the interpolation of the function values stored in this lookup table at the keys that enclose this sample. 
A monotonic constraint for the $i$-th variable is satisfied by enforcing that the interpolation 
parameters of adjacent keypoints follow the desired constraints.
\citet{Gupta2016} formulated this as a convex optimization problem with linear inequality
constraints. These constraints guarantee that the fitted model respects the desired monotonicity.

ALAMO~\citep{cozad2015combined} is a commercial data-driven parametric 
symbolic model that supports monotonicity constraints. 
It uses the BARON~\citep{sahinidis1996baron} global optimization solver
to find optimal parameters for a model with non-linear constraints
representing the desired monotonicity.
These constraints are the partial derivatives of the symbolic 
model applied to a sample of the input points, thus creating constraints where only the model parameters are unknown. The approach is again optimistic.



\citet{aubin2020hard} uses a SOCP formulation to enforce monotonicity, 
convexity, and non-negativity constraints in kernel regression models. 
They also resort to the discretization of the constraints by choosing 
a set of virtual points covering a compact set of sufficient points 
to ensure the satisfaction of these constraints. They tested monotonicity and convexity constraints on a one-dimensional problem. 
They did not analyze the size of the compact set for higher 
dimensions, therefore the scalability of this approach is unknown.


Recent work by~\citet{Bladek2019} on counterexample-driven GP is closely related to our work and has a similar objective as it also incorporate domain knowledge into SR using constraints. They use a satisfiability 
solver to check if each candidate model fulfills shape or symmetry constraints by looking for counter-examples of the desired constraint. 
\citet{kubalik2020symbolic} use multi-objective GP to minimize the approximation 
error of the training data and minimize the constraints for constraint data. 
The constraint data are artificially generated to capture the desired constraint. 
So, for example, for a monotonicity constraint, they generate ordered input data and
check whether the evaluated function is increasing (or decreasing).


In \cite{ourPaper} we introduced shape-constrained symbolic regression and
analyzed single-objective GP with penalization for constraint violations and ITEA with a
feasible-infeasible 
population to deal with shape constraints. The main idea is to use IA to
calculate bounds for model outputs as well as partial derivatives of the model
over input variables.
The results showed that SR models without shape constraints are highly likely to exhibit unexpected behavior, even when using noiseless data.
With shape constraints, all tested algorithms could find feasible solutions
but exhibited slightly worse approximation error. We hypothesized that this
was a consequence of a slower convergence and loss of diversity induced by the
rejection of invalid solutions and the false negatives caused by the imprecision of IA calculations.

  

\section{Methodology}
\label{sec:method}


This paper proposes a multi-objective approach that 
minimizes the model error and the constraint violations (soft-constraints).

Following~\cite{ourPaper}, we use IA (Section~\ref{subsec:intervalArihmetic}) 
to estimate bounds. As previously mentioned, 
IA gives a pessimistic approximation of the true interval, 
thus we have the guarantee that the returned model is feasible.
A comparison with an optimistic approach or 
a hybridization of both will be studied in future work.


\subsection{Single-objective Genetic Programming}

Our implementation of single-objective genetic programming (GP) follows the 
traditional evolutionary meta-heuristic process flow starting with a random
population of symbolic expressions followed by a fitness evaluation, selection
of parents, a recombination of pairs of
expressions, generating a child population, and a mutation operator applied
to these children.
The symbolic expressions are represented as $n$-ary tree with constants 
and variables at the leaves. 
Within the selection step,
individuals from the current generation are selected to serve as parents for the
next generation. We use tournament selection to select the parents. The recombination
(crossover) combines
two subtrees sampled from the parent solutions. The mutation operations can change either a single node or an entire subtree. Finally, the current population is 
replaced by the children and the process is repeated until the 
stop criterion is met.

Optionally memetic local optimization is applied in fitness evaluation~\cite{Kommenda:GPEM}, where a number of number of iterations of non-linear least squares optimization is applied after fitness evaluation in order to improve parameter values in the expression models.

We evaluate the quality of each solution with the normalized mean of squared 
error (NMSE as shown in Equation~\ref{eqn:nmse}). Infeasible solutions are 
assigned a high NMSE value so that they are not selected for recombination.

The algorithm used to evaluate an expression tree with IA is the same as the one used to evaluate the data 
points. The only difference being the use of specific implementation of
operators and functions for IA.
Since we are only using differentiable functions within our function set, we
can calculate the partial derivatives of the expression trees symbolically.



\subsection{Multi-objective Approach}
\label{subsec:multiObjectiveApproach}

We tested two different multi-objective
optimization algorithms: the non-dominated sorting genetic algorithm
(NSGA-II)~\cite{Deb2002} and the multi-objective evolutionary algorithm based on
decomposition (MOEA/D)~\cite{Zhang2007}. It is not our intention to explicitly
compare these two approaches to each other in terms of quality, in fact we expect
to observe different outcomes for different problem instances since we have problems
with a small and with a large number of objectives. Because NSGA-II is a classic
domination-based EA, it usually performs better when optimizing a small number of
objectives.


For both algorithms we used an adaptation for multi-objective symbolic regression
proposed by~\citet{Kommenda2015} to prevent unwanted inflation of
the Pareto-front with solutions which have only minor differences in objective
values.

In both algorithms we handle each constraint as a separate
objective. This results in $1 + n$ objectives for each problem instance, whereas $n$ is
the number of constraints to minimize. The first objective is to minimize the
NMSE. We use soft constraints instead of hard constraints as in the
single-objective approach. Using soft constraints allows us
to minimize the amount of violation for each constraint to get the notion 
of a solution being closer to feasibility than others. The calculation of the
$i$-th objective-function $P_i$ is:

\begin{equation}
    \begin{aligned}
    \label{equation:calcSoftPenalties}
        P_i &= P^{\text{inf}}_i + P^{\text{sup}}_i\\
        P^{\text{inf}}_i &= \lvert \min(\text{inf}(f_i(x)) - \text{inf}(c_i), 0) \rvert \\
        P^{\text{sup}}_i &= \lvert \max(\text{sup}(f_i(x)) - \text{sup}(c_i), 0) \rvert,
    \end{aligned}
\end{equation}
with $f_i(.)$ being the evaluation of the interval corresponding to the
$i$-th constraint, either as the image of the function or one of the partial
derivatives, $c_i$ the feasibility interval for the $i$-th constraint, $\text{inf}(x), \text{sup}(x)$ are
functions that return the inferior and superior bounds of the interval.

Besides the potential benefit of improving evolutionary dynamics for constrained
optimization problems the multi-objective approach also provides a set
of Pareto-optimal solutions. This can be of advantage if some of the constraints
are difficult or impossible to fulfill as the algorithm still produces
partially valid solutions.

\section{Experiments}
%
We analyze single and multi-objective algorithms
in two different scenarios. In the first set of experiments we analyze
the predictive error with different levels of noise for new points
sampled from the same distribution as training points (\emph{in-domain
performance}). For the second set of experiments we evaluate the predictive error for points
outside of the domain of the training points (\emph{out-of-domain
performance}). For the second set of experiments, we extend the set of problem instances and include
three additional problem instances specifically selected to highlight out-of-domain performance.

We report the normalized mean
of squared errors (NMSE in percent) for a hold-out set of data points not used in training. The NMSE for target vector
$y$ and prediction vector
$\hat{y}$ with $N$ elements is defined as in Equation~\ref{eqn:nmse}.
\begin{equation}
  \label{eqn:nmse}
  \text{NMSE}(y, \hat{y}) = \frac{100}{\text{var}(y) N} \sum_{i=1}^{N}{(y_i - \hat{y}_i)^2}
\end{equation}

\section{Analysis of In-domain Performance}
\label{sec:analysisInDomain}
\subsection{Problem Instances}
\label{subsec:problemInstances}
We use a subset of problem instances from the \emph{Feynman Symbolic
Regression Database}~\cite{Feynman} which is a collection of models
from physics textbooks. We select instances based on the
difficulty reported in~\cite{Feynman}, whereby we kept only the harder
instances\footnote{We found that many of the problem instances are
trivial to solve, in particular there are some instances that are only
products of two variables or ratios of two or three variables.} for
which we could easily derive monotonicity constraints.The selected functions are shown in
Table~\ref{table:instanceExpressions}. They are smooth, non-linear,
and have only a few inputs.

For each instance we defined a number of shape constraints derived
from the known expression. The shape constraints are shown in Table
\ref{tab:constraintDefinitions} whereby we use a compact tuple
notation. The first element of the tuple is the allowed range for
model output and the remaining elements indicate for each input
variable whether the partial derivative of the model over that
variable is non-positive ($-1$, monotone decreasing) or non-negative
($1$, monotone increasing). The value zero indicates no constraint on
the partial derivative.

\begin{table}[ht!]
    \center
    \caption{Instance expressions taken from \citep{Feynman}}
    \label{table:instanceExpressions}
    \renewcommand{\arraystretch}{1.5}
    \begin{tabular}{p{90pt}p{155pt}}
        Instance & Expression\\
        \hline
        I.6.20        & $\operatorname{exp}\left( \frac{-{{\left( \frac{\theta}{\sigma}\right) }^{2}}}{2}\right) \frac{1}{\sqrt{2 \pi} \sigma}$\\
        I.9.18        & $\frac{G\, \mathit{m1}\, \mathit{m2}}{{{\left( \mathit{x2}-\mathit{x1}\right) }^{2}}+{{\left( \mathit{y2}-\mathit{y1}\right) }^{2}}+{{\left( \mathit{z2}-\mathit{z1}\right) }^{2}}}$\\
        I.15.3x       & $\frac{x-u t}{\sqrt{1-\frac{{{u}^{2}}}{{{c}^{2}}}}}$\\
        I.30.5        & $\operatorname{asin}\left( \frac{\mathit{lambd}}{n d}\right)$\\
        I.32.17       & $\frac{1}{2} \epsilon c\, {{\mathit{Ef}}^{2}}\, \frac{8 \pi {{r}^{2}}}{3}\, \frac{{{\omega}^{4}}}{{{\left( {{\omega}^{2}}-{{{{\omega}_0}}^{2}}\right) }^{2}}}$\\
        I.41.16       & $\frac{h\, {{\omega}^{3}}}{{{\pi}^{2}}\, {{c}^{2}}\, \left( \operatorname{exp}\left( \frac{h \omega}{\mathit{kb} T}\right) -1\right) }$\\
        I.48.20       & $\frac{m\, {{c}^{2}}}{\sqrt{1-\frac{{{v}^{2}}}{{{c}^{2}}}}}$\\
        II.6.15a      & $\frac{\frac{{p_d}}{4 \pi \epsilon} 3 z}{{{r}^{5}}} \sqrt{{{x}^{2}}+{{y}^{2}}}$\\
        II.11.27      & $\frac{n \alpha}{1-\frac{n \alpha}{3}} \epsilon \mathit{Ef}$\\
        II.11.28      & $1+\frac{n \alpha}{1-\frac{n \alpha}{3}}$\\
        II.35.21      & ${n_{\mathit{rho}}} \mathit{mom} \operatorname{tanh}\left( \frac{\mathit{mom} B}{\mathit{kb} T}\right)$\\
        III.9.52      & $\frac{{p_d} \mathit{Ef} t}{h} {{\sin{\left( \frac{\left( \omega-{{\omega}_0}\right)  t}{2}\right) }}^{2}}$\\
        III.10.19     & $\mathit{mom}\, \sqrt{{{\mathit{Bx}}^{2}}+{{\mathit{By}}^{2}}+{{\mathit{Bz}}^{2}}}$\\
    \end{tabular}
\end{table}

\begin{table*}[ht!]
  \caption{Shape constraints used in each problem instance. The \emph{input
  space} column refers to the variable domains and the \emph{constraints} column
  a tuple with the first element being the allowed range for the model output
  and the remaining elements corresponding to the variables. A value of $1$ represents
  a non-decreasing constraint, $-1$ a non-increasing, $0$ when there is no
  constraint.}
    \label{tab:constraintDefinitions}
    \resizebox{\textwidth}{!}{%
    \begin{tabular}{p{1.7cm}>{$}l<{$}>{$}l<{$}}
        Instance & \text{Input space} & \text{Constraints}\\
        \hline
        I.6.20        & (\sigma, \theta) \in [1..3]^2 & ([0..\infty], 0, -1)\\
        I.9.18        &
        \left(\mathit{x1},\mathit{y1},\mathit{z1},\mathit{m1},\mathit{m2},G,\mathit{x2},\mathit{y2},\mathit{z2}\right) & ([0..\infty], -1, -1, -1, 1, 1, 1, 1, 1, 1)\\
        &\in [3..4]^3\times[1..2]^6&\\
        I.15.3x       & (x, u, t, c) \in [5..10]\times[1..2]^2\times[3..20]& ([0..\infty], 1, 0, -1, -1)\\
        I.30.5        & \left( \mathit{lambd},n,d\right) \in [1..5]^2\times[2..5]& ([0..\infty], 1, -1, -1)\\
        I.32.17       & \left( \epsilon,c,\mathit{Ef},r,\omega,{{\omega}_0}\right) \in [1..2]^5\times[3..5]& ([0..\infty], 1, 1, 1, 1, 1, -1)\\
        I.41.16       & \left( \omega,T,h,\mathit{kb},c\right)  \in [1..5]^5& ([0..\infty], 0, 1, -1, 1, -1)\\
        I.48.20       & (m, v, c) \in [1..5]\times[1..2]\times[3..20]& ([0..\infty], 1, 1, 1)\\
        II.6.15a      & \left( \epsilon,{p_d},r,x,y,z\right) \in [1..3]^6& ([0..\infty], -1, 1, -1, 1, 1, 1)\\
        II.11.27      & \left( n,\alpha,\epsilon,\mathit{Ef}\right) \in [0..1]^2\times[1..2]^2& ([0..\infty], 1, 1, 1, 1)\\
        II.11.28      & \left( n,alpha\right) \in[0..1]^2& ([0..\infty], 1, 1)\\
        II.35.21      & \left( {n_{\mathit{rho}}},\mathit{mom},B,\mathit{kb},T\right)  \in [1..5]^5& ([0..\infty], 1, 1, 1, -1, -1)\\
        III.9.52      & \left( {p_d},\mathit{Ef},t,h,\omega,{{\omega}_0}\right) \in  [1..3]^4\times[1..5]^2 & ([0..\infty], 1, 1, 0, -1, 0, 0)\\
        III.10.19     & \left( \mathit{mom},\mathit{Bx},\mathit{By},\mathit{Bz}\right)  \in [1..5]^4 & ([0..\infty], 1, 1, 1, 1)\\
    \end{tabular}}

\end{table*}

\subsection{Data Sampling and Partitioning}
We sample data points uniformly at random from the input space shown
in Table~\ref{tab:constraintDefinitions}. 100 points are assigned to
the training partition, 100 points to a validation set for grid search and 100 points are assigned to the test
partition.
The relatively
small data size in combination with high noise levels decreases the
chance of finding well-fitting models.

\subsection{Simulated Noise}
Similarly to our previous work \cite{ourPaper} we generate four
variations of each problem instance with different noise
levels. However, we use much higher noise levels than previously. The first version contains the target variable without noise. The other three versions
are obtained by adding different levels of noise to the target variables $y' = y +
N(0, \sqrt{x}\sigma_{y})$, where $x$ specifies the noise level. For the purpose
of this experiment, we have tested noise levels of $x = \{10\%, 30\%, 100\%\}$.
Noise is only added to the training and validation data so the optimally
achievable NMSE on the test set is $0\%$ for all problem instances and noise levels.

%


\subsection{Experiment Setup}
\label{subsec:experimentsetup}
For each instance we execute $30$ independent runs for each of the GP algorithms. 
%
First, we use the algorithms without shape
constraints. These results serve as baseline for the achievable
prediction error and are also indicative of the capability of GP to find
feasible models even without shape constraints.

The same experiments are performed with the extended algorithms
including shape constraints. The aim is to verify whether feasible models
can be found more easily using shape constraints and to compare the NMSE to the results without
shape constraints.


\subsection{Algorithm Configuration}
\label{subsec:algorithmConfiguration}
We use single- and multi-objective algorithms for the experiments:
single-objective tree-based genetic programming (GP), GP with memetic local
optimization of model parameters as described in~\cite{Kommenda:GPEM} (GPOpt),
GP using shape constraints (GPSC), GP using shape constraints and local memetic
optimization (GPOptSC),
multi-objective tree-based GP based on the non-dominated sorting genetic
algorithm II (NSGA-II) and the multi-objective
evolutionary algorithm based on decomposition (MOEA/D). The multi-objective
algorithms do not include memetic local optimization.

For each algorithm we run a grid search to tune
parameter settings.
For the common parameters for GP, GPOpt, GPSC, GPOptSC, NSGA-II and
MOEA/D, we vary the maximal tree
size from 10 to 50 nodes with an increment of 10. For the function
set we used the following options: $\text{F}_1, \text{F}_2, \text{F}_3 \text{ and
} \text{F}_4$ as shown in Table \ref{table:common-parameters}. We use the configuration with the
smallest NMSE on the validation set to re-train the final model on the training set.

\begin{table}[ht!]
    \caption{Parameter settings for GP, GPOpt, NSGA-II, MOEA/D}
    \center
    \label{table:common-parameters}
    \setlength{\tabcolsep}{3pt}
    \resizebox{\textwidth}{!}{%
    \begin{tabular}{ll}
    Parameters      & Value\\
    \hline
        Function set            & $\text{F}_1 = \{+, \times, -\}$\\
                                & $\text{F}_2 = \text{F}_1 \cup \{AQ(x, y)\}$\\
                                & $\text{F}_3 = \text{F}_2 \cup \{\sqrt{x}, x^2,
                                \log, \exp\}$\\
                                & $\text{F}_4 = \text{F}_3 \cup \{\sin, \tanh\}$\\                       
        Terminal set            & parameters, weight * variable\\
        Max. tree depth         & 20 levels\\
        Max. tree length        & $\{10, 20, 30, 40, 50\}$ nodes\\
        Tree initialization     & Probabilistic tree creator (PTC2)\\
        Max. evaluated solutions& 500000\\
        Population size         & 1000 individuals\\
        Generations             & 500\\
                                & 50 with local optimization of parameters\\
        Selection               & Tournament selection with group size 5\\
        Crossover               & Subtree crossover\\
        Crossover probability   & 100\%\\
        Mutation Operators      & Select one of the following randomly:\\
                                & \tabitem Replace subtree with random branch\\
                                & \tabitem Change a function symbol\\
                                & \tabitem Add $x \sim N(0, 1)$ to all numeric parameters\\
                                & \tabitem Add $x \sim N(0, 1)$ to one numeric parameter\\
        Mutation rate           & 15\%\\
        Primary objective       & Minimize NMSE\\
        Secondary (multiple) objectives     & Minimize constraint violations (Eq.~\ref{equation:calcSoftPenalties})\\ \\
    \end{tabular}}
\end{table}

The results of evolutionary algorithms are compared to shape-constrained polynomial regression (SCPR) (see
e.g. \cite{hall2018optimization}) which performed well in our earlier experiments \cite{ourPaper}. The implementation relies
on sum\--of\--squares\--programm\-ing (SOS) to incorporate non-negativity
constraints whereby the relaxed optimization problem is formulated as a semidefinite
programming (SDP) problem (see e.g. \cite{parrilo2000structured}). The resulting
SDP problem is solved using the Mosek solver\footnote{https://www.mosek.com}.
The polynomial models were fit using a regularized least-squares
approach using an elastic-net penalty\cite{glmnet}. The first
parameter is the total degree of the polynomial model (e.g. for a
bi-variate problem and total degree of three we have the basis
functions: $1,x,x^2,x^3,y,y^2,y^3,xy,x^2y,xy^2$). The elastic-net parameters are: $\alpha$ to balance between
L1-penalty and L2-penalty, and $\lambda$ to balance between the error
term and the penalty terms. The three parameters were tuned using a
grid-search with 5-fold cross-validation. The degree was varied in the range $[1, 2, \ldots, 6]$, $\alpha \in
\{0, 0.5, 1\}$, and $\lambda \in
\{1\cdot10^{-9},1\cdot10^{-8},1\cdot10^{-7},1\cdot10^{-6},1\cdot10^{-5},1\cdot10^{-4},0.001,
0.01, 0.1, 1.0\}$. The same grid-search was executed for each problem
instance. For each instance the best cross-validated MSE was
determined and then from all configurations with similar results
(CV-MSE not worse than one standard deviation from the best) the
configuration with smallest degree, largest lambda and largest
alpha was selected. This configuration was used to train the final
model on the whole training set. The same procedure was used for
polynomial regression (PR) and shape-constrained polynomial regression (SCPR).



%

\subsection{Results: Constraint Violations}
\label{subsec:constraintViolations}

\begin{figure}[H]
    \centering
    \includegraphics[scale=0.50]{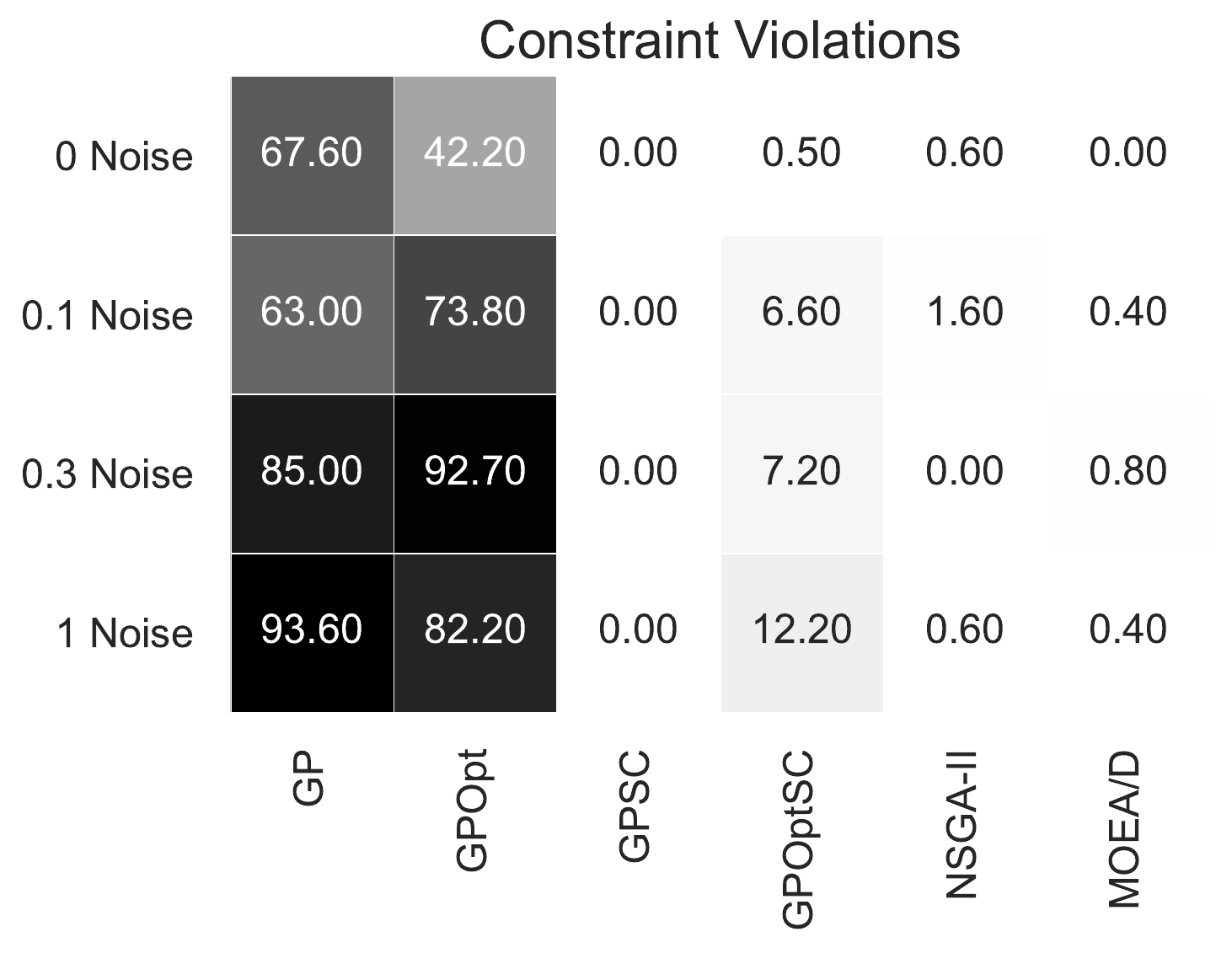}
    \caption{Percentage of infeasible models generated by each algorithm over 30 runs of all instances.}
    \label{fig:constraintViolations}
\end{figure}

Figure \ref{fig:constraintViolations} shows the percentage of infeasible
solutions across all instances broken down by noise level. To calculate the
percentage of violation we counted all the solutions which violate at least one
constraint and divide it by the amount of models over all instances. For the feasibility
test we uniformly sample 100000 points of the input space for each instance and
take them to generate a feasibility holding dataset. Then we evaluate our final SR
solutions as well as the partial derivatives against this dataset. A solution is infeasible when at least one constraint is
violated. Figure \ref{fig:constraintViolations} shows that the algorithms without shape constraints have a high
probability to produce infeasible solutions. Only on easier
problem instances with no or low noise some of the identified solutions fulfill all
the constraints. This is the case when the generating function is re-discovered.
With shape constraints the probability to identify feasible solutions is much
higher. This is of course expected as the first group of algorithms is not aware
of the constraints. Especially for the GPSC approach where infeasible solution
candidates are rejected, we always find solutions which conform to the expected
behavior. GPOptSC turns out worse than GPSC, as we apply parameter
optimization after the evaluation, which can lead to constraint violations, due
to a linear scaling applied to the solutions after the feasibility check. Here we have additionally tested multi-objective algorithms.
The results in Figure \ref{fig:constraintViolations} show that NSGA-II and MOEA/D have a similar probability to identify
feasible solutions as the single-objective algorithms. Even though we do not reject infeasible solutions in the
multi-objective approach, the algorithms succeed in evolving feasible solutions
with a high probability.

\subsection{Results: Sensitivity to Noise}
\label{subsec:impactOfnoise}
\label{subsec:qualityAnalysis}
Tables \ref{tab:resultNMSELow} and \ref{tab:resultNMSEHigh} show the median NMSE of the best solution for each
instance out of 30 runs. Table \ref{tab:resultNMSELow} shows results for no noise and $10\%$ noise, Table \ref{tab:resultNMSEHigh} shows results for $30\%$ and $100\%$ noise. 

The result tables are split in two sections. The section on the left side shows the
error values of the algorithms without shape constraints. The section on the
right side shows results with shape constraints. The best result in each row is highlighted.

Analyzing the tables we notice that algorithms with shape constraints are not
better on instances with no or low noise. This is consistent with the
finding in~\cite{ourPaper}. However, here we observe that with increasing noise shape constraints help to improve the results.
The row counts where the algorithms with shape constraints are better/equal/worse than without shape constraints are $(2, 6, 5)$ without noise, $(12, 0, 1)$ with 10\% noise, $(9, 0, 4)$ with 30\% noise, and $(10, 0, 3)$ with 100\% noise. 
\begin{table*}[!ht]
    \caption{Median NMSE values for test with no noise and 10\% noise}
    \centering
    \resizebox{\textwidth}{!}{%
    \label{tab:resultNMSELow}

    \begin{tabular}[]{c@{}l>{$}r<{$}>{$}r<{$}>{$}r<{$}|>{$}r<{$}>{$}r<{$}>{$}r<{$}>{$}r<{$}>{$}r<{$}>{$}r<{$}@{}}
        & & \multicolumn{3}{c |}{w/o. info} & \multicolumn{5}{c}{w. info}\\
        & & \text{GP} & \text{GPOpt} & \text{PR} & \text{GPSC} &
        \text{GPOptSC} & \text{SCPR} & \text{NSGA-II} & \text{MOEA/D}\\
        \hline
        \parbox[t]{1.5em}{\multirow{19}{*}{\rotatebox[origin=c]{90}{no noise}}}
        & I.6.20       & 0.31 & \textbf{0.00} & 0.01 & 2.25 & \textbf{0.00} & 0.16 & 0.10 & 0.32 \\
        & I.9.18       & 0.98 & \textbf{0.60} & 1.05 & 1.12 & 1.21 & 1.20 & 0.81 & 1.20 \\
        & I.15.3x      & 0.08 & \textbf{0.00} & 0.06 & 0.09 & 0.01 & 0.08 & 0.03 & 0.03 \\
        & I.30.5       & \textbf{0.00} & \textbf{0.00} & 0.14 & 0.15 & \textbf{0.00} & 0.25 & \textbf{0.00} & \textbf{0.00} \\
        & I.32.17      & 0.04 & 0.06 & 50.70 & 0.66 & 0.34 & 7.64 & \textbf{0.02} & 0.80 \\
        & I.41.16      & \textbf{0.18} & 0.26 & 9.10 & 5.61 & 5.20 & 7.51 & 3.78 & 8.03 \\
        & I.48.20      & \textbf{0.00} & \textbf{0.00} & \textbf{0.00} & \textbf{0.00} & \textbf{0.00} & \textbf{0.00} & \textbf{0.00} & \textbf{0.00} \\
        & II.6.15a     & 0.46 & \textbf{0.01} & 17.15 & 1.43 & 0.61 & 27.63 & 0.13 & 0.43 \\
        & II.11.27     & \textbf{0.00} & \textbf{0.00} & 0.02 & \textbf{0.00} & \textbf{0.00} & 0.02 & \textbf{0.00} & \textbf{0.00} \\
        & II.11.28     & \textbf{0.00} & \textbf{0.00} & 0.00 & \textbf{0.00} & \textbf{0.00} & \textbf{0.00} & \textbf{0.00} & \textbf{0.00} \\
        & II.35.21     & 0.53 & \textbf{0.00} & 0.96 & 2.96 & 0.46 & 1.45 & 0.75 & 3.90 \\
        & III.9.52     & 11.47 & 10.14 & 55.56 & 71.79 & 8.51 & 43.08 & \textbf{7.41} & 41.36 \\
        & III.10.19    & 0.29 & \textbf{0.00} & \textbf{0.00} & 0.62 & 0.02 & \textbf{0.00} & 0.07 & 0.74 \\
        \hline
        \parbox[t]{1.5em}{\multirow{19}{*}{\rotatebox[origin=c]{90}{noise 10\%}}}
        & I.6.20       & 2.00 & 1.83 & 2.73 & 3.30 & \textbf{1.53} & 3.17 & 2.54 & 2.96 \\
        & I.9.18       & 2.92 & 5.34 & 6.82 & 2.82 & 3.98 & 4.62 & \textbf{2.20} & 3.42 \\
        & I.15.3x      & 1.59 & 2.05 & 3.17 & 1.37 & 1.79 & 5.14 & \textbf{0.95} & 1.38 \\
        & I.30.5       & \textbf{0.13} & 0.32 & 4.75 & 0.92 & 1.03 & 2.61 & 0.52 & 0.87 \\
        & I.32.17      & 2.56 & 4.45 & 32.48 & 2.57 & 5.15 & 9.29 & \textbf{1.97} & 4.85 \\
        & I.41.16      & 3.89 & 5.79 & 23.19 & 5.98 & 7.24 & 12.27 & \textbf{3.87} & 8.17 \\
        & I.48.20      & 0.87 & 1.13 & 6.67 & \textbf{0.31} & \textbf{0.31} & 4.11 & \textbf{0.31} & \textbf{0.31} \\
        & II.6.15a     & 3.43 & 7.51 & 26.26 & 3.09 & 2.32 & 40.66 & \textbf{1.32} & 7.65 \\
        & II.11.27     & 0.72 & 0.60 & 3.30 & 0.71 & 0.64 & 3.30 & \textbf{0.53} & 0.73 \\
        & II.11.28     & 0.28 & 0.60 & 0.98 & \textbf{0.13} & \textbf{0.13} & 1.16 & \textbf{0.13} & 0.41 \\
        & II.35.21     & 2.33 & 2.14 & 3.60 & 4.07 & 3.93 & 12.36 & \textbf{1.88} & 5.26 \\
        & III.9.52     & 15.06 & 11.09 & 59.90 & 78.43 & 31.38 & 53.41 & \textbf{7.56} & 85.22 \\
        & III.10.19    & 1.70 & 1.29 & 3.48 & 1.66 & 1.42 & 6.29 & \textbf{1.22} & 1.78 \\
    \end{tabular}}
\end{table*}

\begin{table*}[!ht]
    \caption{Median NMSE values for test with 30\% and 100\% noise}
    \centering
    \resizebox{\textwidth}{!}{%
    \label{tab:resultNMSEHigh}
    \begin{tabular}[]{c@{}l>{$}r<{$}>{$}r<{$}>{$}r<{$}|>{$}r<{$}>{$}r<{$}>{$}r<{$}>{$}r<{$}>{$}r<{$}>{$}r<{$}@{}}
        & & \multicolumn{3}{c |}{w/o. info} & \multicolumn{5}{c}{w. info}\\
        & & \text{GP} & \text{GPOpt} & \text{PR} & \text{GPSC} &
        \text{GPOptSC} & \text{SCPR} & \text{NSGA-II} & \text{MOEA/D}\\
        \hline
        \parbox[t]{1.5em}{\multirow{19}{*}{\rotatebox[origin=c]{90}{noise 30\%}}}
        & I.6.20       & 5.33 & 4.15 & 5.22 & 4.31 & 4.24 & 7.28 & \textbf{2.61} & 3.27 \\
        & I.9.18       & 1.89 & 3.42 & 7.63 & 1.65 & 2.28 & 3.78 & \textbf{1.05} & 2.58 \\
        & I.15.3x      & 3.71 & 3.54 & \textbf{1.86} & 5.16 & 2.66 & 4.30 & 2.25 & 4.15 \\
        & I.30.5       & 2.77 & \textbf{2.53} & 11.52 & 3.34 & 3.71 &8.89 & 3.24 & 3.32 \\
        & I.32.17      & 4.53 & 13.26 & 78.01 & 4.90 & 5.69 & 47.90 & \textbf{3.33} & 5.35 \\
        & I.41.16      & 10.00 & 13.08 & 56.95 & 8.50 & 9.81 & 62.72 & \textbf{4.73} & 12.20 \\
        & I.48.20      & 1.94 & 2.18 & 5.73 & \textbf{1.07} & \textbf{1.07} & 3.24 & 2.10 & 1.99 \\
        & II.6.15a     & 7.66 & 10.32 & 40.73 & 3.13 & 15.07 & 50.17 & \textbf{2.68} & 11.37\\
        & II.11.27     & 6.26 & 6.61 & 26.94 & 3.12 & \textbf{2.70} & 11.47 & 2.76 & 3.01 \\
        & II.11.28     & 2.44 & 2.44 & \textbf{1.48} & 1.78 & 1.78 & 1.57 & 1.78 & 2.46 \\
        & II.35.21     & 4.76 & 6.14 & 16.05 & 4.43 & 4.08 & 5.07 & \textbf{3.10} & 4.76 \\
        & III.9.52     & 39.53 & 16.83 & 56.24 & 74.84 & 39.10 & 52.13 & \textbf{12.23} & 75.67 \\
        & III.10.19    & \textbf{4.24} & \textbf{4.24} & 6.42 & 4.68 & 4.76 & 4.33 & 4.79 & 4.69 \\
        \hline
        \parbox[t]{1.5em}{\multirow{19}{*}{\rotatebox[origin=c]{90}{noise 100\%}}}
        & I.6.20       & 4.28 & 4.31 & 56.27 & 5.67 & 5.82 & 56.26 & \textbf{4.06} & 6.08 \\
        & I.9.18       & 9.03 & 19.08 & 9.02 & 9.21 & 18.54 & 28.37 & \textbf{6.51} & 18.91 \\
        & I.15.3x      & 6.69 & 12.60 & 12.61 & 7.07 & 7.28 & 12.61 & \textbf{6.17} & 6.81 \\
        & I.30.5       & \textbf{1.97} & 11.39 & 15.77 & 6.01 & 8.71 & 4.99 & 6.56 & 5.98 \\
        & I.32.17      & 24.76 & 46.56 & 108.27 & 9.59 & 12.31 & 69.67 & \textbf{7.30} & 14.55 \\
        & I.41.16      & 27.57 & 55.76 & 102.81 & 23.86 & 27.03 & 102.81 & \textbf{19.24} & 30.77 \\
        & I.48.20      & 3.75 & 4.89 & 13.45 & 4.38 & 4.13 & \textbf{2.59} & 4.93 & 4.25 \\
        & II.6.15a     & 24.22 & 29.62 & 100.18 & 17.01 & 18.77 & 82.35 & \textbf{16.59} & 21.89\\
        & II.11.27     & \textbf{2.19} & 22.52 & 106.57 & 6.11 & 5.88 & 52.93 & 3.04 & 5.24 \\
        & II.11.28     & 2.72 & 2.67 & 23.42 & \textbf{2.54} & 2.59 & 23.42 & 2.61 & 2.64 \\
        & II.35.21     & 23.45 & 19.25 & 80.97 & 17.66 & 17.52 & 80.97 & \textbf{17.10} & 18.34 \\
        & III.9.52     & 65.65 & 45.94 & 100.21 & 82.67 & 78.29 & 100.21 & \textbf{26.62} & 84.44 \\
        & III.10.19    & 9.20 & \textbf{5.41} & 53.01 & 7.01 & 11.64 & 12.03 & 7.75 & 7.63 \\
    \end{tabular}}
\end{table*}


\begin{figure}
    \begin{subfigure}[b]{0.5\textwidth}
        \includegraphics[width=\textwidth]{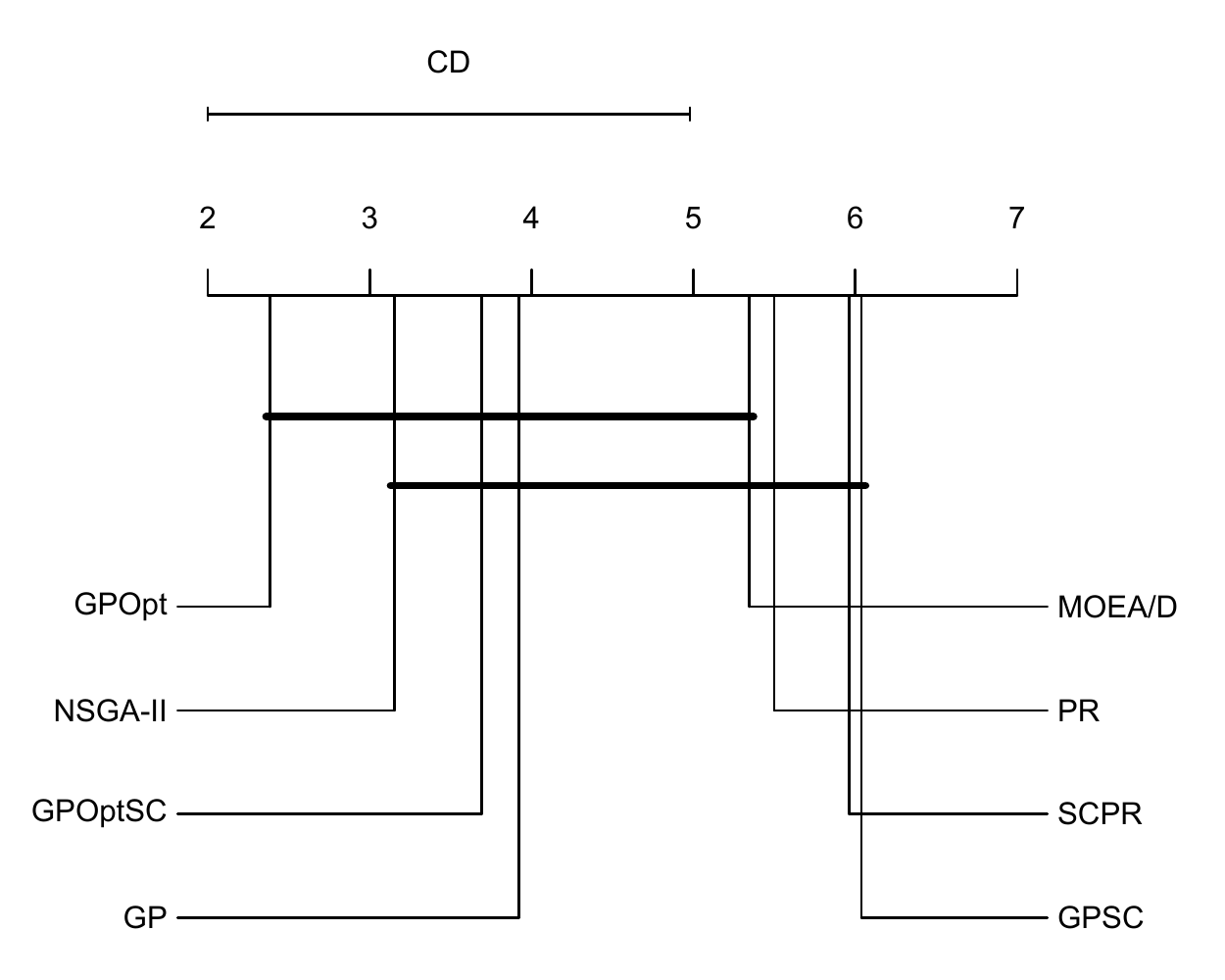} 
        \caption{CD plot on noise level 0\%}
        \label{fig:cd_plot_noise_0}
    \end{subfigure}
    \begin{subfigure}[b]{0.5\textwidth}
        \includegraphics[width=\textwidth]{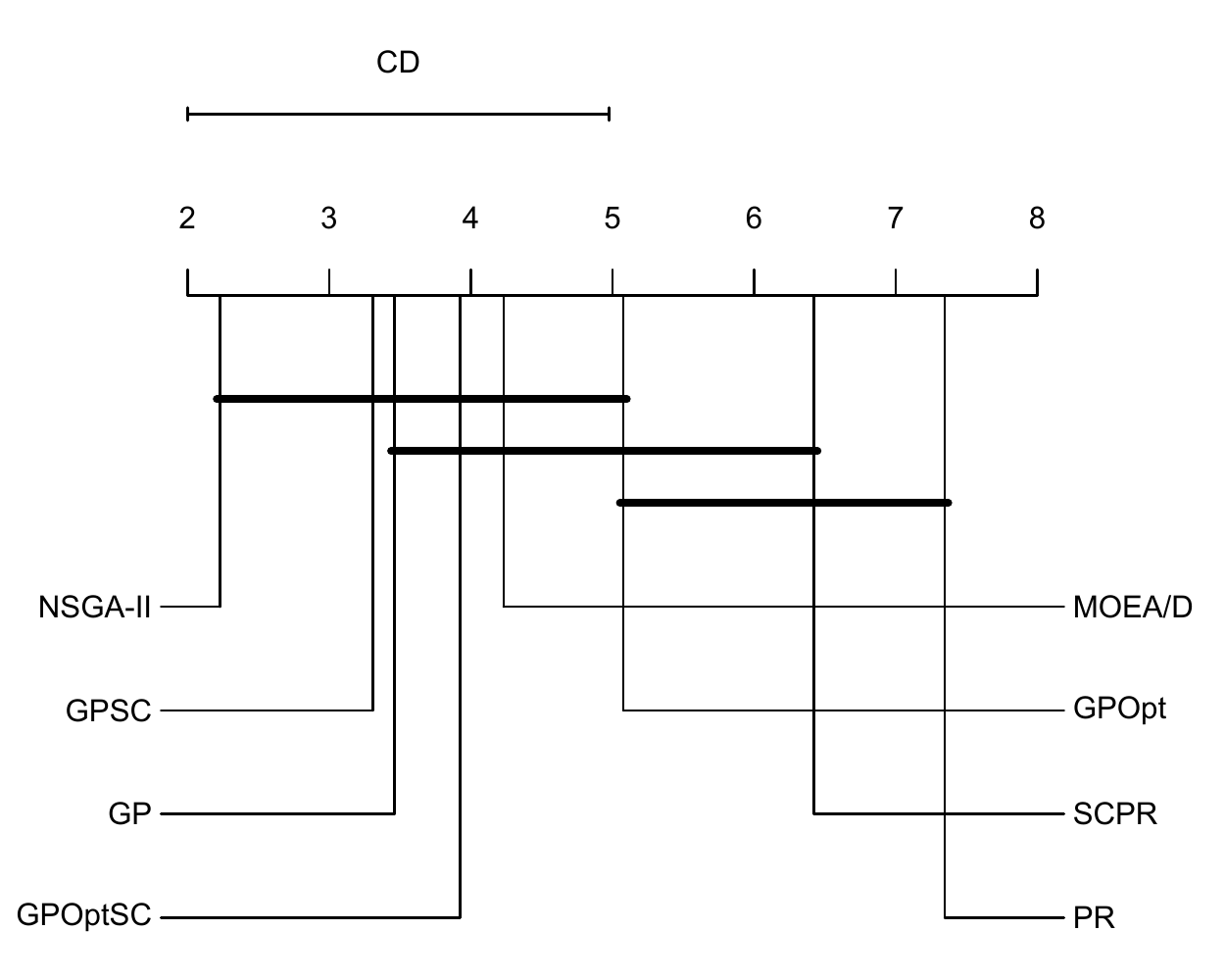}
        \caption{CD plot on noise level 100\%}
        \label{fig:cd_plot_noise_1}
    \end{subfigure}
    \caption{Critical difference plots in-domain performance}
    \label{fig:cd_plots_noise}
\end{figure}

The critical difference (CD) plots in Figure \ref{fig:cd_plots_noise} show the
average rank of the different algorithms. As we can see
in Figure \ref{fig:cd_plot_noise_0} GPOpt has the best overall rank and is significantly
better than PR, SCPR and GPSC. All other algorithms do not have
a significant difference to each other and GPSC performs worst without noise. In
Figure \ref{fig:cd_plot_noise_1} the CD plot for highest noise level shows, that the algorithms with shape
constraints rank better than for the instances without noise.
For the highest noise level NSGA-II performs best overall and
GPSC has the second best rank.



PR and SCPR do not
perform well compared to the evolutionary algorithms. The CD plots in Figure \ref{fig:cd_plots_noise} show that PR and
SCPR rank significant worse than the best algorithms. This is in
contrast to \cite{ourPaper} where we observed that SCPR produced the best results for no noise or
low-noise problem instances. The reason for this difference is that in
the new experiments we tuned hyper-parameters for the evolutionary
algorithms using a grid search and achieved better prediction errors. At the same
time hyper-parameter selection for PR and SCPR was changed to
improve results for high-noise problem instances which lead to
increased prediction errors for the low noise experiments.

\section{Analysis of Out-of-domain Performance}
\label{sec:analysisOutOfDomain}
The results above show that shape constraints can help to find SR models that
conform to expected behavior especially for high noise settings. 
However, we did not detect a significant difference to the best algorithms
without shape constraints.
Shape constraints may also improve
the prediction error for extrapolation when new observations lie outside
of the domain of the training error. In this second set of experiments we
try to find evidence for this hypothesis.

\subsection{Problem instances}
For analysing the out-of-domain prediction errors, we use three
additional problem instances: \emph{Kotanchek} and 
\emph{UnwrappedBall} from~\cite{vladislavleva2009order}, and a
problem instance from~\cite{pagie1997evoluationary} in the following
called \emph{Pagie}. The first two problem instances are chosen because they were also used
for analysing extrapolation capabilities of SR in \cite{vladislavleva2009order}
and additionally because they are monotonic in large subsets of the input space.

Table \ref{tab:out-of-domain-instanceExpressions} shows the generating
expressions for the three additional problem instances, Table
\ref{tab:out-of-domain-constraintDefinitions} shows the definition of
the input space and the shape constraints. For these three instances
we use different constraints on non-overlapping subspaces of the input
space whereby we split the input spaces at the extrema of the three
functions which are located at $(0,0)$ for \emph{Pagie}, $(1,2.5)$ for
\emph{Kotanchek} and $(3,3,3,3,3)$ for \emph{UnwrappedBall}.

For the extrapolation instances, we generate samples within the input domain
until the training, validation and test partitions had $100$ samples
each. To split the data into training and test sets we use the condition that
all samples that lie in the first $x\%$ or the last $x\%$ of the domain range,
are assigned to the test set. For the \emph{Feynman}, \emph{Kotanchek} and
\emph{UnwrappedBall} instances we use $10\%$ for the test set and for \emph{Pagie} we
increase the fraction to $30\%$. This procedure ensures that the test samples
are located in the outer hull of the data set. We add noise to the
target values in the training sets as described above. 


\begin{table}[ht!]
    \center
    \caption{Expressions for the additional problem instances}
    \label{tab:out-of-domain-instanceExpressions}
    \renewcommand{\arraystretch}{1.5}
    \begin{tabular}{p{120pt}p{155pt}}
        Instance & Expression\\
        \hline
        Pagie~\citep{pagie1997evoluationary} & $\frac{1}{1+{{x}^{-4}}} + \frac{1}{1+{{y}^{-4}}}$\\
        Kotanchek~\citep{vladislavleva2009order}  & $\frac{\operatorname{exp}\left( (x1 - 1)^{2}\right)}{1.2 + (x2 - 2.5)^{2}}$\\
        UnwrappedBall~\citep{vladislavleva2009order}  & $\frac{10}{5+\sum_{n=1}^5 (x_{i}-3)^{2}}$\\
    \end{tabular}
\end{table}

\begin{table*}[ht!]
  \caption{Shape constraints used for additional problem instances taken
  from~\citep{pagie1997evoluationary} and \citep{vladislavleva2009order}}
    \label{tab:out-of-domain-constraintDefinitions}
    \resizebox{\textwidth}{!}{%
    \begin{tabular}{p{2.7cm}>{$}l<{$}>{$}l<{$}}
        Instance & \text{Input space} & \text{Constraints}\\
        \hline
        Pagie         & \left( x, y\right) \in  [-5\ldots5]^2 & f(x,y) \in [0..2]\\
                      &                                   & \frac{\partial}{\partial x} f(x,y) \leq 0, x < 0\\
                      &                                   & \frac{\partial}{\partial x} f(x,y) \geq 0, x > 0\\
                      &                                   & \frac{\partial}{\partial y} f(x,y) \leq 0, y < 0\\
                      &                                   & \frac{\partial}{\partial y} f(x,y) \geq 0, y > 0\\ 
        Kotanchek     & \left( x_1, x_2\right) \in [-0.2\ldots4.2]^2 & f(x_1, x_2) \in [0\ldots1]\\
                      &                                         & \frac{\partial}{\partial x_1} f(x_1,x_2) \geq 0, x_1 < 1\\
                      &                                         & \frac{\partial}{\partial x_1} f(x_1,x_2) \leq 0, x_1 > 1\\
                      &                                         & \frac{\partial}{\partial x_2} f(x_1,x_2) \geq 0, x_2 < 2.5\\
                      &                                         & \frac{\partial}{\partial x_2} f(x_1,x_2) \leq 0, x_2 > 2.5\\ 
        UnwrappedBall & \left( x_1, x_2, x_3, x_4, x_5\right)                   & f(\textbf{x}) \in [0\ldots2]\\
                      & \in [-0.25\ldots6.35]^5                                 & \forall_{i\in\{1\ldots5\}}:\\
                      &                                                         & \frac{\partial}{\partial x_i} f(\textbf{x}) \geq 0, x_i < 3\\
                      &                                                         & \frac{\partial}{\partial x_i} f(\textbf{x}) \leq 0, x_i > 3\\ 
    \end{tabular}}    
\end{table*}

\subsection{Experimental setup}
We use the same algorithms and parameter settings as for the first
set of experiments. We re-run the grid-search for the best
configuration because in this set of experiments the input space for
the training set is smaller than before and may have an effect on the
optimal parameter settings. Again 30 independent runs are executed
for each problem instance using the best settings from the grid-search.

\subsection{Results: Effect on Extrapolation}
%
\label{subsec:extrapolation_demo}

\begin{figure}  
    \begin{subfigure}[b]{0.5\textwidth}
    \includegraphics[width=\textwidth]{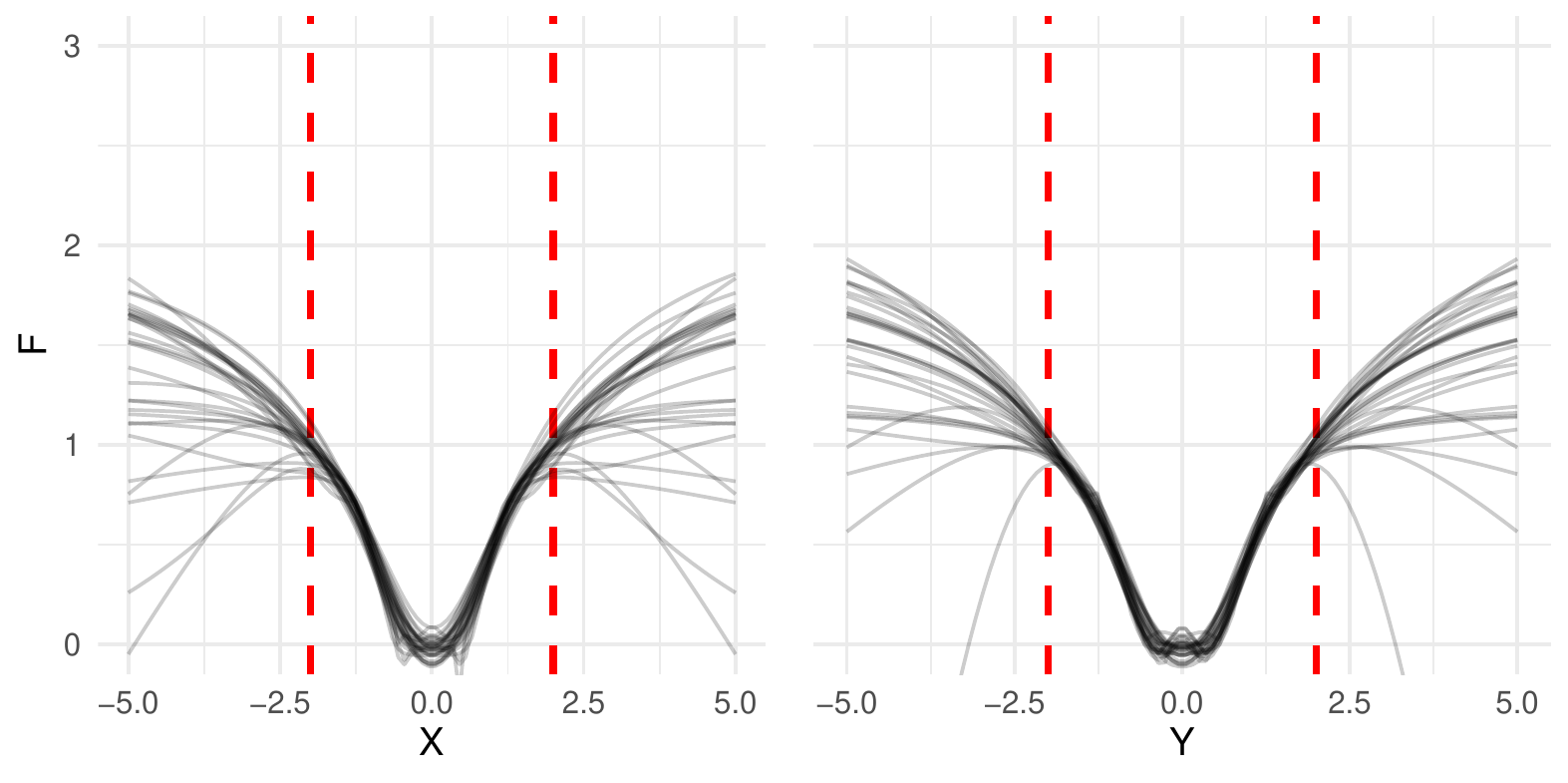}
    \caption{\label{fig:gp-extrapolation}GP}    
    \end{subfigure}
    \begin{subfigure}[b]{0.5\textwidth}
    \includegraphics[width=\textwidth]{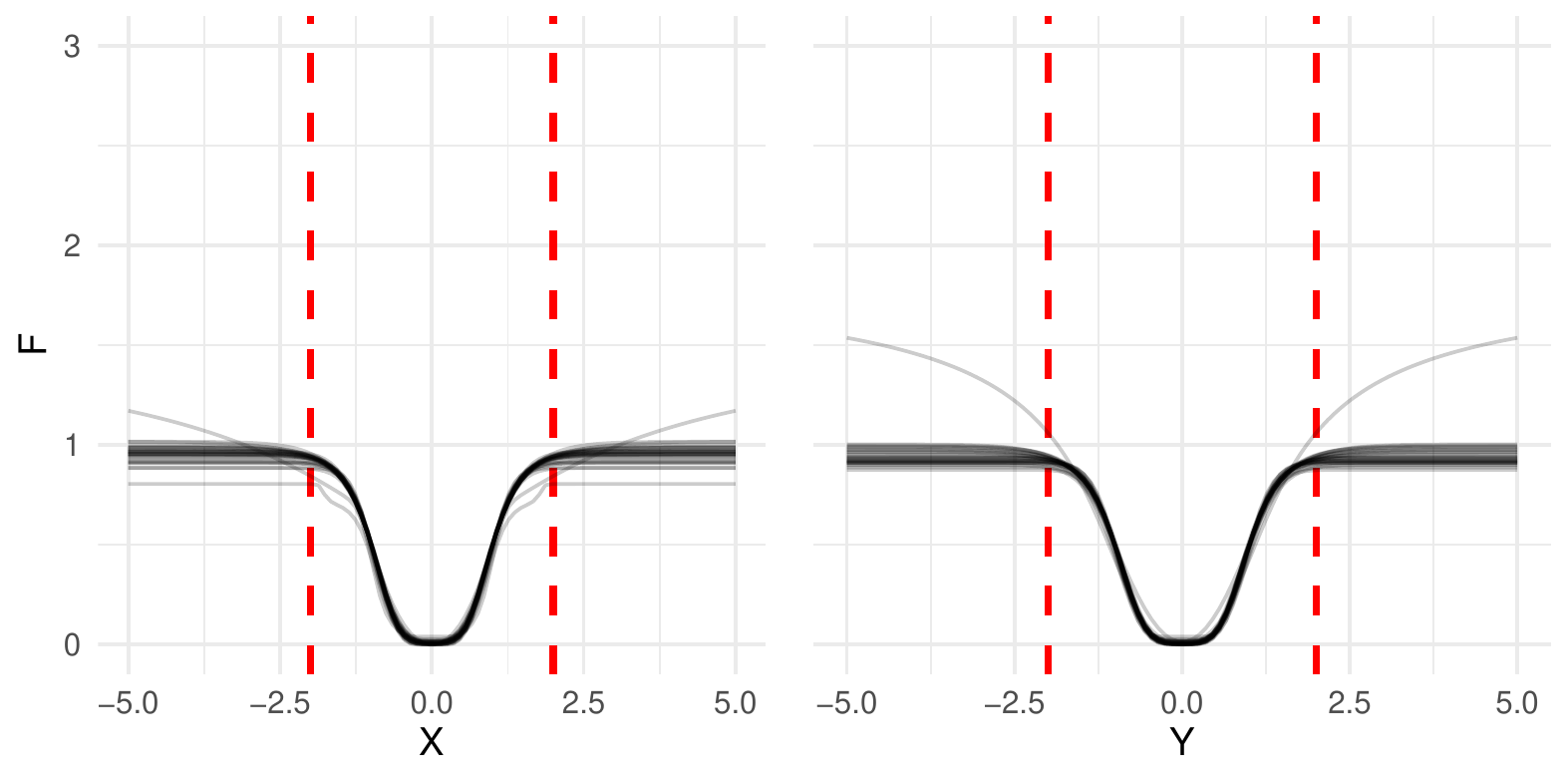}
    \caption{\label{fig:gpsc-extrapolation}GPSC}
    \end{subfigure}
    \vspace{0.3em}

    \begin{subfigure}[b]{0.5\textwidth}
        \includegraphics[width=\textwidth]{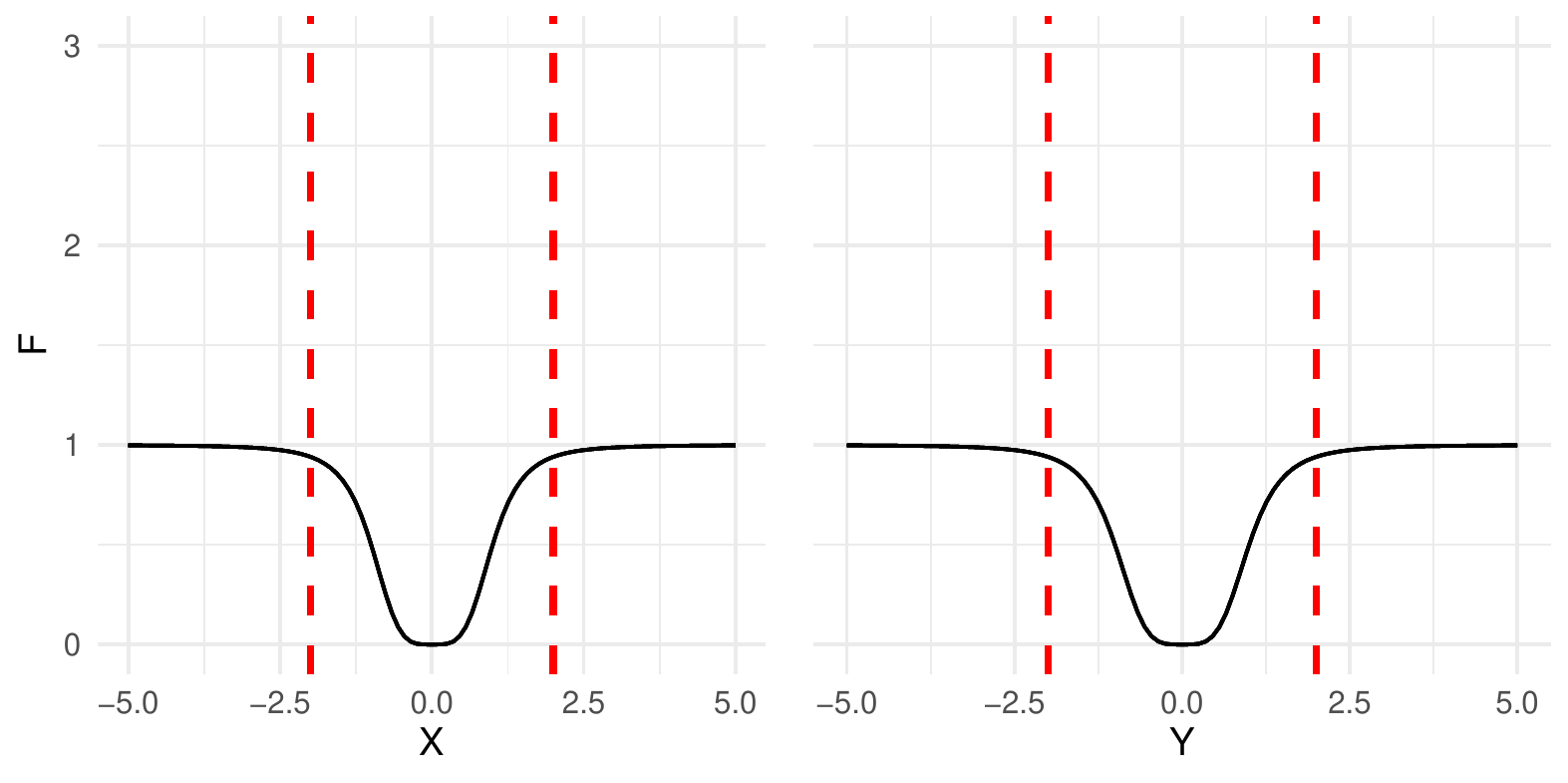}
        \caption{\label{fig:gpopt-extrapolation}GPOpt}        
        \end{subfigure}
        \begin{subfigure}[b]{0.5\textwidth}
        \includegraphics[width=\textwidth]{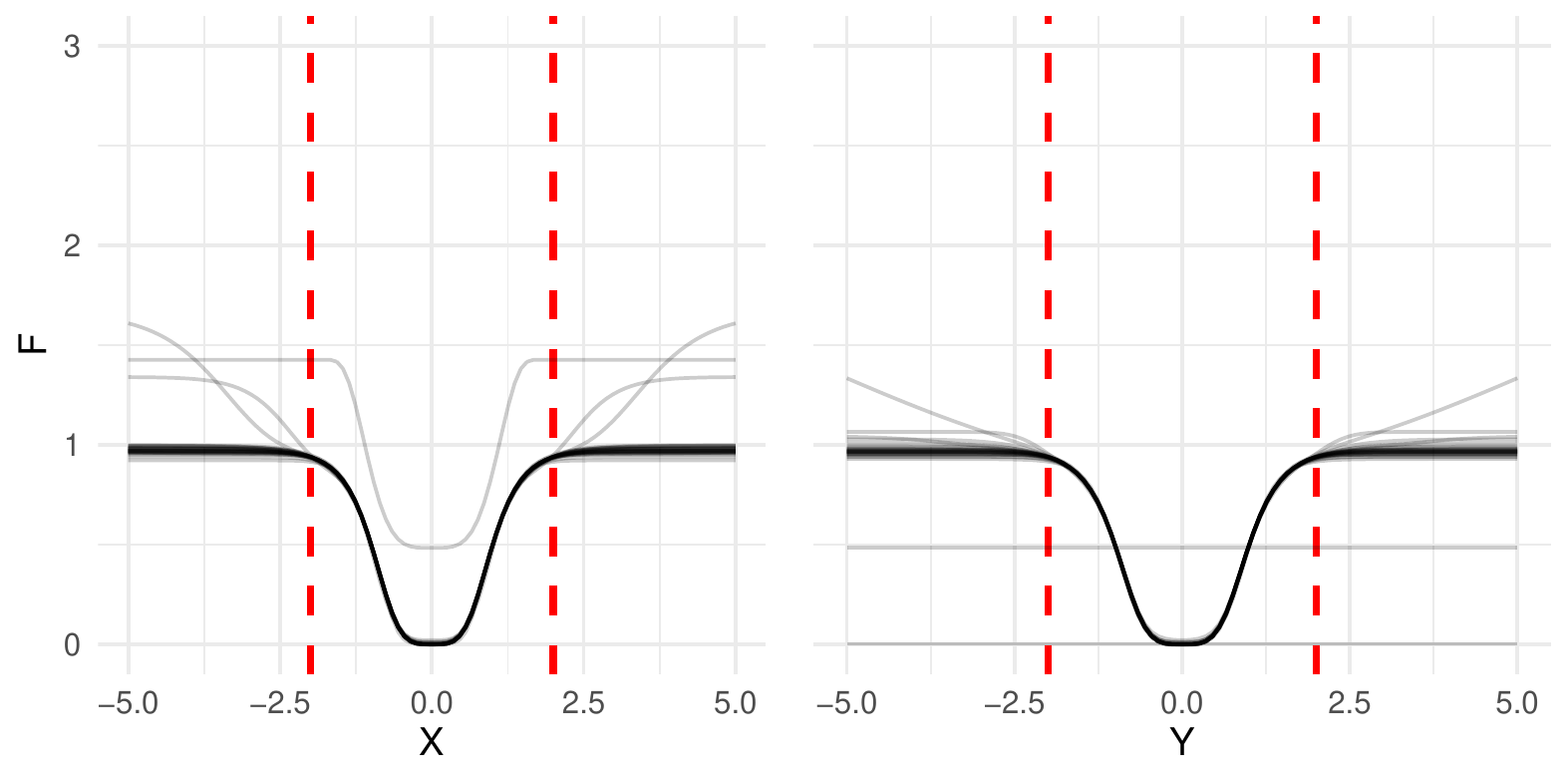}
        \caption{\label{fig:gpoptsc-extrapolation}GPOptSc}        
     \end{subfigure}

    \caption{\label{fig:spatialCo_extrapolation}Partial dependence plots for the 30 models identified for the \emph{Pagie} problem without noise.}
\end{figure}

\begin{figure}  
    \begin{subfigure}[b]{0.5\textwidth}
    \includegraphics[width=\textwidth]{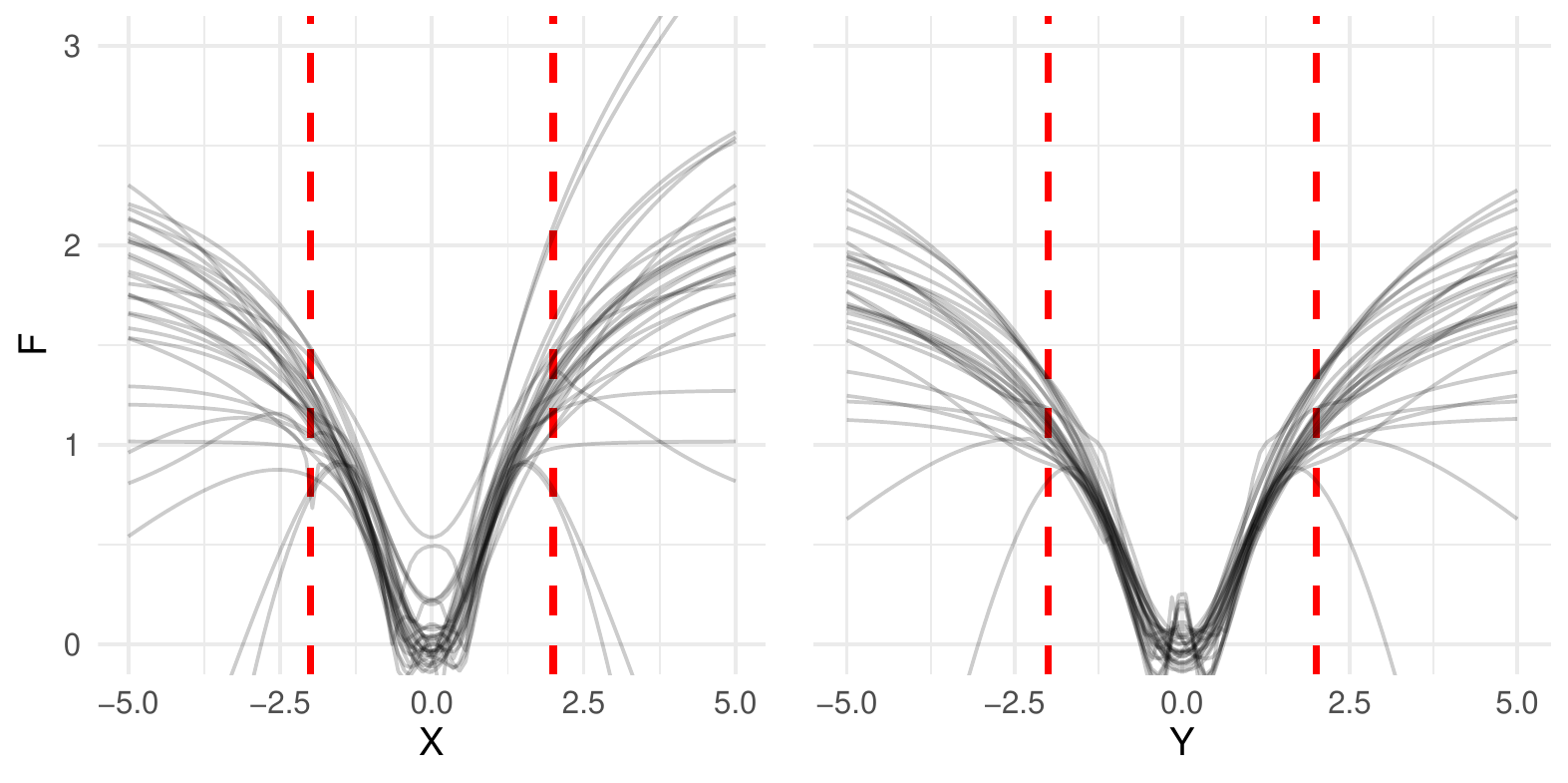}
    \caption{GP}
    \label{fig:gp}
    \end{subfigure}
    \begin{subfigure}[b]{0.5\textwidth}
    \includegraphics[width=\textwidth]{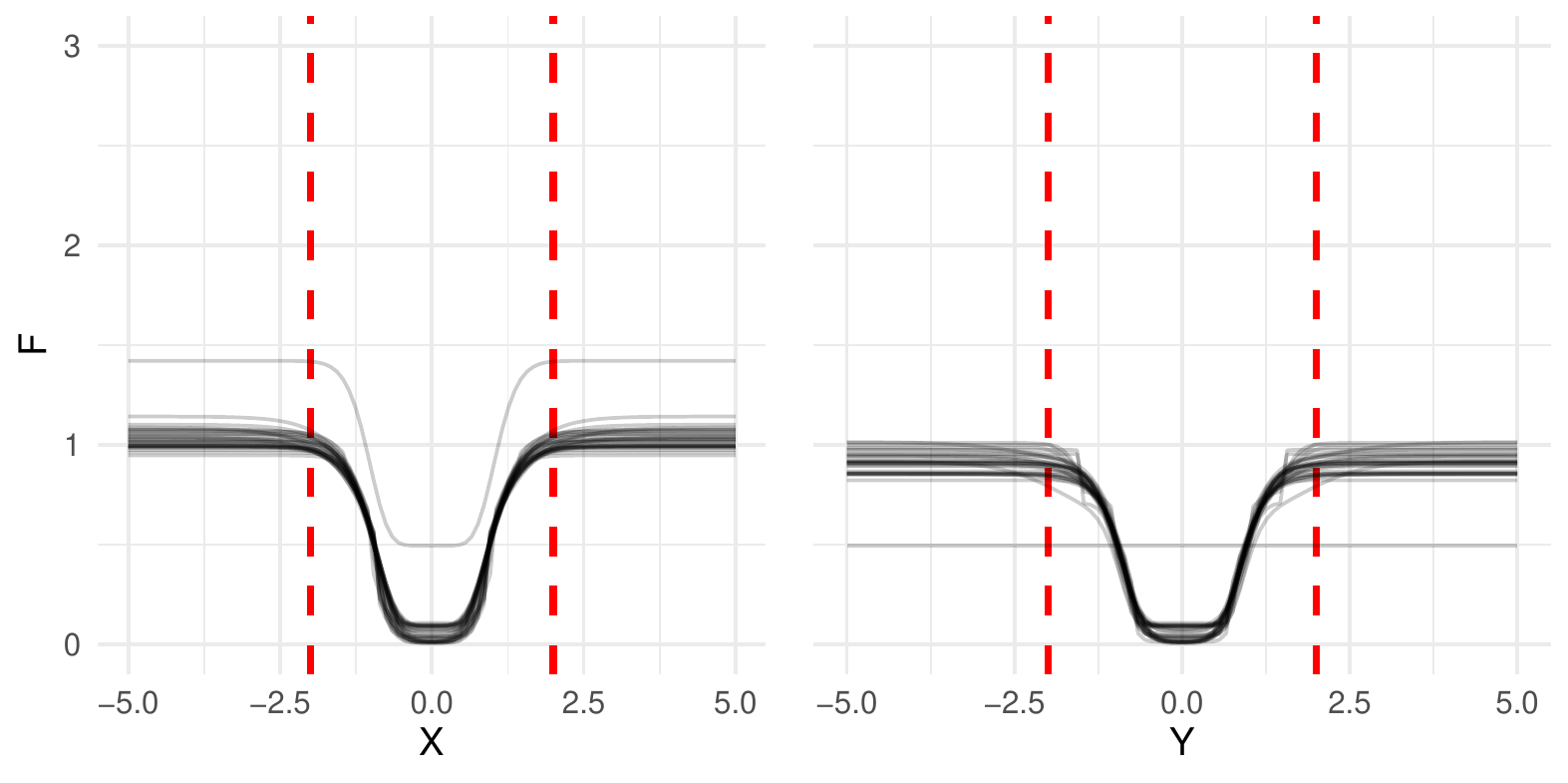}
    \caption{GPSC}
    \label{fig:gpsc}
    \end{subfigure}
    \vspace{0.3em}

    \begin{subfigure}[b]{0.5\textwidth}
        \includegraphics[width=\textwidth]{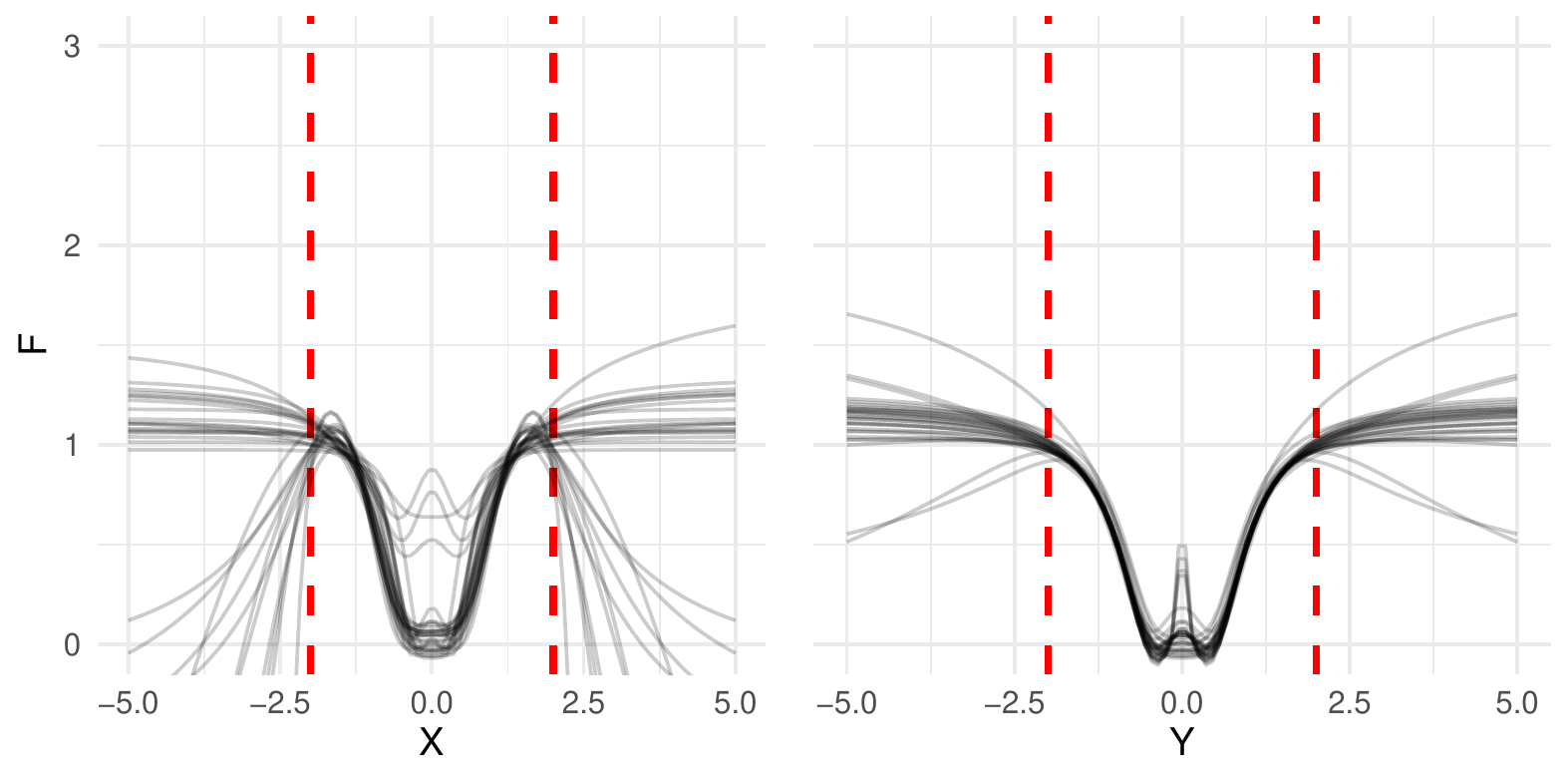}
        \caption{GPOpt}
        \label{fig:gpopt}
        \end{subfigure}
        \begin{subfigure}[b]{0.5\textwidth}
        \includegraphics[width=\textwidth]{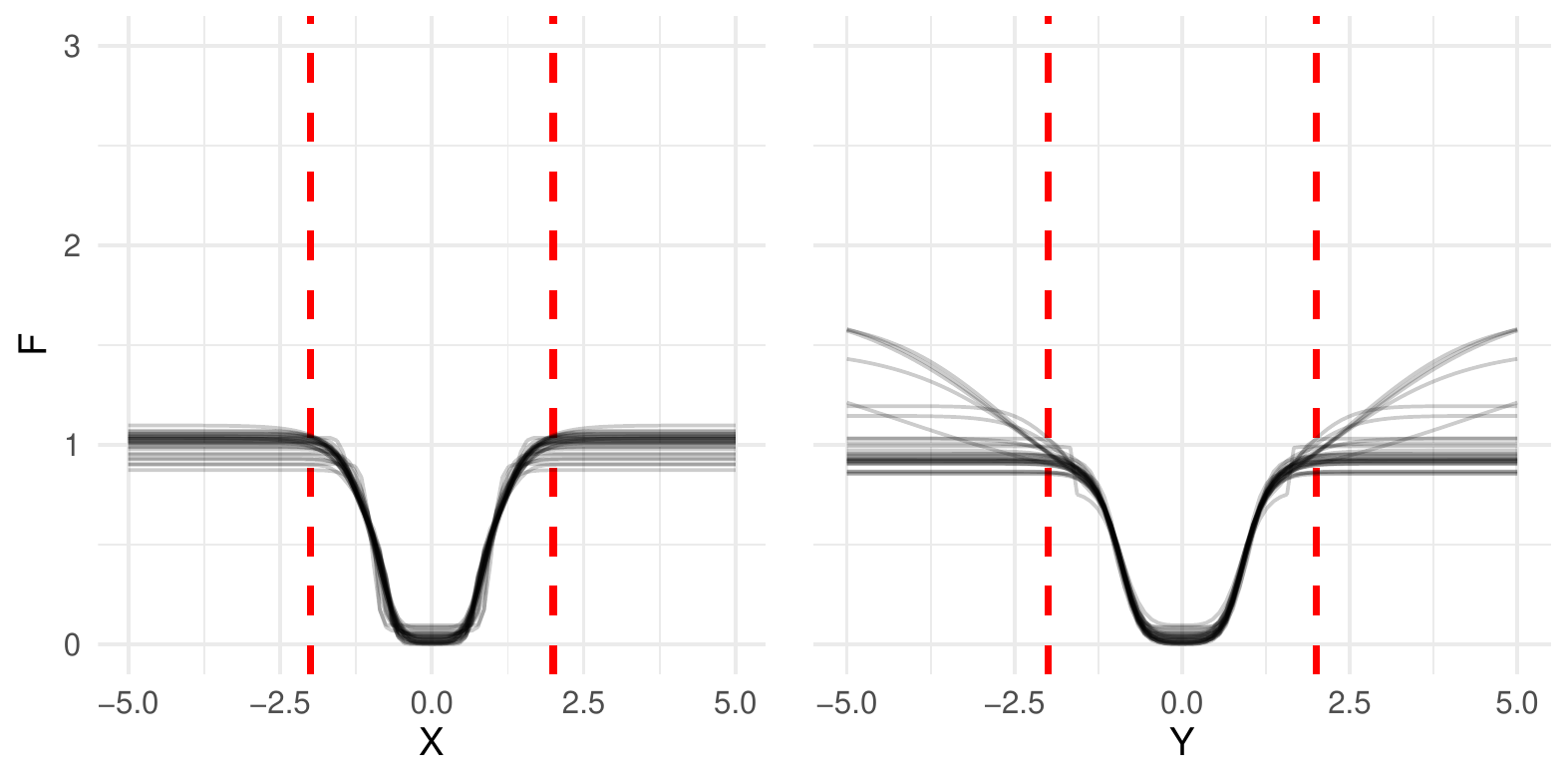}
        \caption{GPOptSc}
        \label{fig:gpoptsc}
        \end{subfigure}

    \caption{\label{fig:spatialCo_extrapolation_0.1}Partial dependence plots for the models identified for the \emph{Pagie} problem with 10\% noise.}
\end{figure}

Figures \ref{fig:spatialCo_extrapolation} and
\ref{fig:spatialCo_extrapolation_0.1} show the effect of shape
constraints using the \emph{Pagie} problem as an example.
The four panels show partial dependence plots for GP and GPOpt and their extensions with shape constraints
(GPSC and GPOptSC). Each partial dependence plot shows the outputs of models found in 30 independent runs over x (with $y=0$) and over y (with $x=0$).

Figure \ref{fig:spatialCo_extrapolation} shows the results without
noise, here GPOpt shows best performance as it identified the
optimal solution in all runs and correspondingly the extrapolation is
perfect even without shape constraints. GP produces the worst models
whereby the algorithm has high variance especially for the extrapolation
region. The introduction of shape constraints into GP significantly improves the
models and produces better
predictions for the extrapolation region. However, for GPOptSC the results
are slightly worse which indicates that the introduction of shape constraints
makes it harder to identify optimal solutions.

Figure \ref{fig:spatialCo_extrapolation_0.1} shows the same plots with
$10\%$ noise. GP again produces the worst models and has high variance
in the extrapolation region. GPSC is again much better and produces
very similar results as without noise. With noise, GPOpt is not able to
identify the optimal solution anymore and tends to overfit, leading to
extreme extrapolation (especially over input variable X). GPOptSC is again much better than GPOpt and similar to GPSC.

These results highlight that shape
constraints can improve prediction errors for noisy problems
especially for extrapolation and that overfitting can be reduced. However, when it is possible to reliably find the optimal
solution then shape constraints may hamper the search.

\begin{figure*}[]
    \centering
    \includegraphics[width=\textwidth]{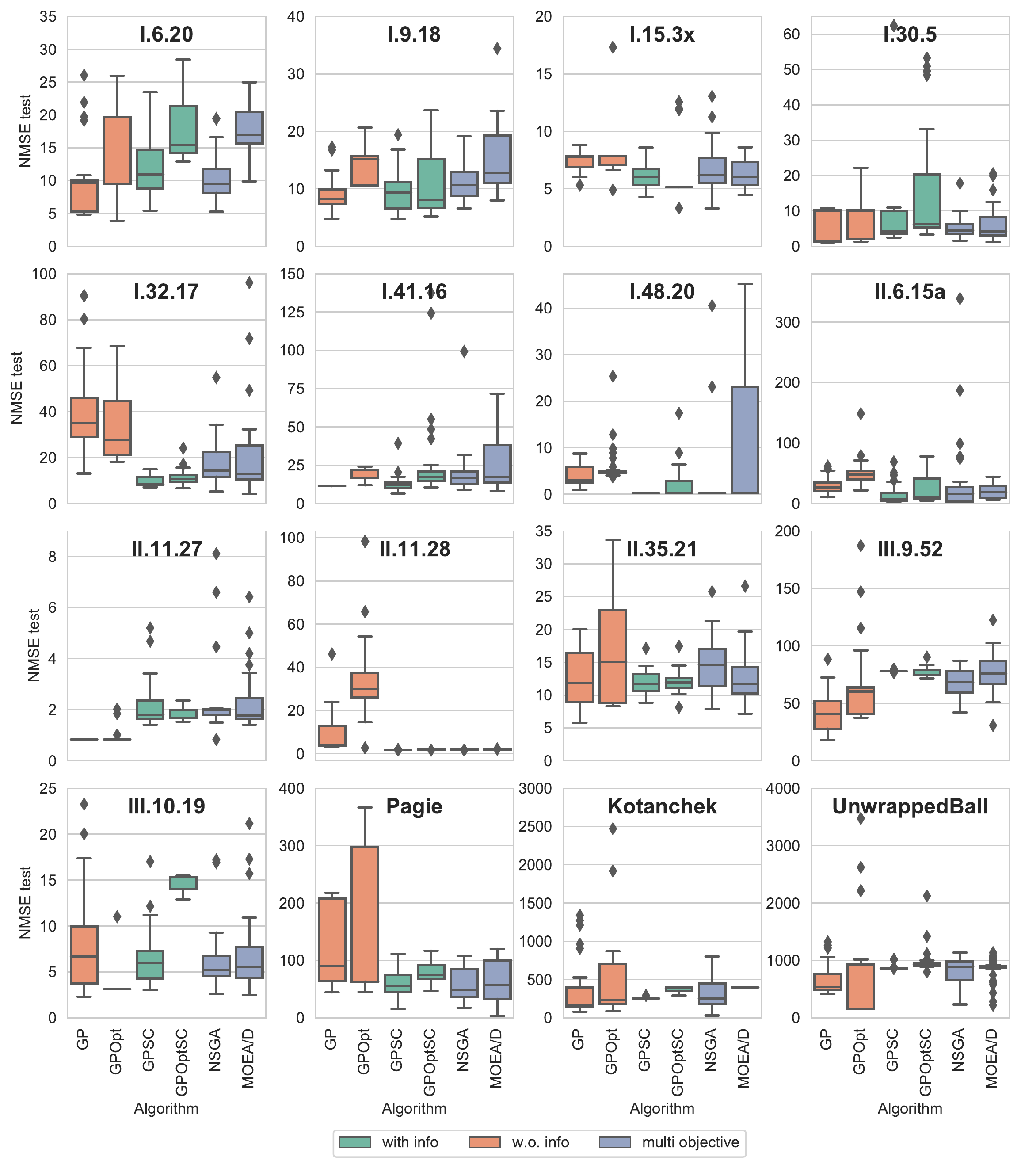}
    \caption[]{NMSE of extrapolation datasets with high noise. The optimal
    achievable NMSE is zero. Following outliers are cut off: I.6.20: GPOpt(1); I.15.3x: NSGA-II(1); I.30.5: NSGA-II(1); I.32.17: NSGA-II(1); I.41.16: GPOptSC(1); II.6.15a: NSGA-II(1), GPOptSC(2); II.35.21: NSGA-II(1); III.9.52: GP(3); Pagie: GPOpt(2); Kotanchek: GPOptSC(4), GPOpt(2); UnwrappedBall: GPOptSC(1), GPOpt(2)}
    \label{fig:nmseBoxplotExtrapolationNoise1}
\end{figure*}

Figure \ref{fig:nmseBoxplotExtrapolationNoise1} shows Box plots over all
extrapolation instances with 100\% noise. On the y-axis the NMSE is
plotted and on the x-axis the different algorithms are plotted. For better
visualization we have limited the range of the y-axis for some plots and cut off
a few outliers. We can also see that there is no algorithm which performs best
over all instances.

Table \ref{tab:resultExtrapolationNMSELow} shows the NMSE over all instances on
the extrapolation dataset for the no noise and $10\%$ noise level. Table
\ref{tab:resultExtrapolationNMSEHigh} shows the high noise levels $30\%$ and
$100\%$. Each of the tables is separated into two section. The left section
shows the error values of the algorithms without using shape constraints and the
right section shows the algorithms with shape constraints. The best result for
each row is highlighted.

\begin{table*}[!ht]
    \caption{Median NMSE values for extrapolation datasets with no noise and 10\% noise }
    \centering
    \resizebox{\textwidth}{!}{%
    \label{tab:resultExtrapolationNMSELow}
    \begin{tabular}[]{c@{}l>{$}r<{$}>{$}r<{$}>{$}r<{$}|>{$}r<{$}>{$}r<{$}>{$}r<{$}>{$}r<{$}>{$}r<{$}>{$}r<{$}@{}}
        & & \multicolumn{3}{c |}{w/o. info} & \multicolumn{5}{c}{w. info}\\
        & & \text{GP} & \text{GPOpt} & \text{PR} & \text{GPSC} &
        \text{GPOptSc} & \text{SCPR} & \text{NSGA-II} & \text{MOEA/D}\\
        \hline
        \parbox[t]{1.5em}{\multirow{12}{*}{\rotatebox[origin=c]{90}{no noise}}}
        & I.6.20      & 1.30 & \textbf{0.00} & 0.08 & 7.20 & 0.32 & 0.04 & 3.49 & 11.26\\
        & I.9.18      & 0.91 & \textbf{0.50} & 0.81 & 17.78 & 0.79 & 0.62 & 1.68 & 1.54\\
        & I.15.3x     & 0.18 & \textbf{0.00} & 0.04 & 0.27 & 0.03 & 0.04 & 0.27 & 0.42\\
        & I.30.5      & 0.01 & \textbf{0.00} & 0.41 & 0.39 & 0.05 & 0.53 & 0.81 & 1.32\\
        & I.32.17     & 0.13 & \textbf{0.05} & 12.79 & 0.74 & 0.50 & 7.14 & 1.62 & 1.52\\
        & I.41.16     & 1.44 & \textbf{0.25} & 1.94 & 7.74 & 4.58 & 2.78 & 6.23 & 6.94\\
        & I.48.20     & \textbf{0.00} & \textbf{0.00} & \textbf{0.00} & \textbf{0.00} & \textbf{0.00} & \textbf{0.00} & \textbf{0.00} & \textbf{0.00}\\
        & II.6.15a    & 1.07 & \textbf{0.19} & 27.03 & 5.19 & 1.73 & 14.26 & 2.11 & 7.75\\
        & II.11.27    & \textbf{0.00} & \textbf{0.00} & \textbf{0.00} & \textbf{0.00} & \textbf{0.00} & \textbf{0.00} & \textbf{0.00} & \textbf{0.00}\\
        & II.11.28    & \textbf{0.00} & \textbf{0.00} & \textbf{0.00} & \textbf{0.00} & \textbf{0.00} & \textbf{0.00} & \textbf{0.00} & \textbf{0.00}\\
        & II.35.21    & 3.10 & \textbf{0.01} & 1.63 & 5.44 & 1.44 & 1.28 & 4.45 & 7.18\\
        & III.9.52    & 7.34 & \textbf{0.48} & 194.32 & 44.67 & 22.25 & 194.37 & 50.40 & 61.42\\
        & III.10.19   & 0.38 & \textbf{0.00} & \textbf{0.00} & 0.71 & 0.02 & \textbf{0.00} & 0.76 & 0.75\\
        & Pagie   & 47.49 & \textbf{0.00} & 4.62e8 & 3.91 & 1.06 & 64.36 & 3.01 & 3.75\\
        & Kotanchek      & 28.47 & \textbf{0.00} & 2.72e5 & 265.58 & 266.22 & 6.60 & 266.22 & 266.22\\
        & UnwrappedBall      & 185.11 & \textbf{25.02} & 919.27 & 804.25 & 895.51 & 775.13 & 805.21 & 807.69\\
        \hline
        \parbox[t]{1.5em}{\multirow{12}{*}{\rotatebox[origin=c]{90}{noise 10\%}}}
        & I.6.20      & 2.17 & \textbf{1.54} & 4.81 & 5.75 & 6.29 & 3.85 & 4.23 & 12.10\\
        & I.9.18      & 28.24 & 26.92 & 9.12 & 17.78 & 13.85 & 6.08 & \textbf{2.73} & 3.40\\
        & I.15.3x     & 1.49 & \textbf{1.47} & 1.99 & 1.50 & 1.49 & 1.99 & 1.49 & 1.58\\
        & I.30.5      & \textbf{0.31} & 0.57 & 8.79 & 1.53 & 1.40 & 7.04 & 0.99 & 1.29\\
        & I.32.17     & \textbf{1.85} & 5.36 & 29.39 & 3.39 & 4.17 & 17.19 & 3.07 & 3.22\\
        & I.41.16     & \textbf{3.62} & 9.04 & 25.61 & 5.32 & 7.24 & 5.64 & 5.53 & 7.41\\
        & I.48.20     & 0.32 & 1.35 & 2.63 & \textbf{0.13} & \textbf{0.13} & 2.00 & 2.00 & \textbf{0.13}\\
        & II.6.15a    & \textbf{1.99} & 8.22 & 62.56 & 4.04 & 5.02 & 40.89 & 2.74 & 7.45\\
        & II.11.27    & 0.19 & 0.61 & 4.89 & \textbf{0.10} & 0.20 & 2.41 & 0.11 & 0.11\\
        & II.11.28    & 0.92 & 1.97 & 2.72 & \textbf{0.02} & \textbf{0.02} & 3.44 & \textbf{0.02} & \textbf{0.02}\\
        & II.35.21    & \textbf{5.42} & 5.80 & 9.10 & 9.11 & 5.80 & 7.51 & 7.83 & 9.87\\
        & III.9.52    & \textbf{8.23} & 9.58 & 89.44 & 50.50 & 22.70 & 75.46 & 33.08 & 62.23 \\
        & III.10.19   & 0.92 & 0.98 & 2.05 & 0.83 & \textbf{0.86} & 2.06 & 0.87 & 0.87\\
        & Pagie   & 79.11 & 32.41 & 3.88e5 & 20.44 & 22.52 & 63.47 & \textbf{18.63} & 20.01\\
        & Kotanchek      & 38.63 & 15.19 & 91.83 & 265.25 & 265.25 & \textbf{14.46} & 265.25 & 265.25\\
        & UnwrappedBall       & 219.3 & \textbf{78.2} & 105.99 & 836.42 & 919.98 & 561.10 & 821.39 & 843.12\\
    \end{tabular}}
\end{table*}

\begin{table*}[!ht]
    \caption{Median NMSE values for extrapolation datasets with 30\% noise and 100\% noise }
    \centering
    \resizebox{\textwidth}{!}{%
    \label{tab:resultExtrapolationNMSEHigh}
    \begin{tabular}[]{c@{}l>{$}r<{$}>{$}r<{$}>{$}r<{$}|>{$}r<{$}>{$}r<{$}>{$}r<{$}>{$}r<{$}>{$}r<{$}@{}}
        & & \multicolumn{3}{c |}{w/o. info} & \multicolumn{5}{c}{w. info}\\
        & & \text{GP} & \text{GPOpt} & \text{PR} & \text{GPSC} &
        \text{GPOptSC} & \text{SCPR} & \text{NSGA-II} & \text{MOEA/D}\\
        \hline
        \parbox[t]{1.5em}{\multirow{12}{*}{\rotatebox[origin=c]{90}{noise 30\%}}}
        & I.6.20      & \textbf{8.95} & 9.56 & 13.41 & 10.42 & 10.65 & 12.68 & 10.03 & 12.23\\
        & I.9.18      & 29.23 & 14.32 & 11.02 & 17.95 & 15.58 & 8.95 & \textbf{5.95} & 6.16\\
        & I.15.3x     & 5.64 & 5.00 & \textbf{1.44} & 4.39 & 1.79 & \textbf{1.44} & 2.88 & 3.25\\
        & I.30.5      & \textbf{1.20} & 5.72 & 20.36 & 3.84 & 5.31 & 12.29 & 4.14 & 3.33\\
        & I.32.17     & \textbf{5.45} & 20.88 & 46.22 & 7.91 & 7.64 & 24.09 & 8.47 & 6.44\\
        & I.41.16     & \textbf{2.94} & 14.21 & 29.83 & 7.69 & 11.61 & 12.31 & 9.87 & 10.07\\
        & I.48.20     & 2.97 & 3.57 & 16.66 & \textbf{0.81} & \textbf{0.81} & 16.32 & \textbf{0.81} & \textbf{0.81}\\
        & II.6.15a    & 12.60 & 14.71 & 60.99 & 5.32 & 13.63 & 48.23 & \textbf{2.87} & 6.63\\
        & II.11.27    & \textbf{1.94} & \textbf{1.94} & 6.89 & 2.10 & 2.05 & 5.39 & 2.19 & 1.96\\
        & II.11.28    & 2.22 & 4.08 & 3.68 & 0.46 & 0.40 & 4.10 & \textbf{0.37} & \textbf{0.37}\\
        & II.35.21    & \textbf{8.34} & 9.69 & 10.18 & 9.41 & 9.21 & 14.86 & 9.15 & 9.33\\
        & III.9.52    & \textbf{13.25} & 15.77 & 67.50 & 51.96 & 42.88 & 46.88 & 54.98 & 69.48\\
        & III.10.19   & 6.12 & \textbf{3.81} & 8.72 & 4.97 & 6.01 & 3.47 & 4.88 & 5.20\\
        & Pagie   & 377.15 & 90.07 & 2.44e5 & \textbf{9.44} & 13.92 & 60.48 & 9.56 & 10.21\\
        & Kotanchek      & 96.11 & \textbf{12.39} & 608.18 & 287.85 & 287.85 & 16.71 & 287.85 & 287.85\\
        & UnwrappedBall       & 385.89 & \textbf{82.75} & 208.41 & 724.45 & 835.37 & 431.90 & 821.01 & 825.21\\
        \hline
        \parbox[t]{1.5em}{\multirow{12}{*}{\rotatebox[origin=c]{90}{noise 100\%}}}
        & I.6.20      & 9.59 & 19.71 & 30.36 & 10.95 & 15.45 & 30.36 & \textbf{8.87} & 17.02\\
        & I.9.18      & 14.13 & 15.16 & 21.23 & 13.19 & 13.79 & 19.34 & \textbf{12.69} & 12.72\\
        & I.15.3x     & 6.91 & 7.06 & 48.39 & 6.05 & 5.13 & 48.39 & \textbf{5.41} & 6.03\\
        & I.30.5      & \textbf{1.47} & 2.04 & 40.17 & 4.35 & 6.15 & 52.04 & 4.66 & 4.11\\
        & I.32.17     & 15.89 & 27.8 & 100.01 & \textbf{8.38} & 10.61 & 59.07 & 11.47 & 12.90\\
        & I.41.16     & \textbf{11.27} & 22.01 & 58.91 & 12.05 & 17.46 & 28.68 & 16.23 & 16.12\\
        & I.48.20     & 2.95 & 4.64 & 14.74 & \textbf{0.25} & \textbf{0.25} & 27.65 & \textbf{0.25} & \textbf{0.25}\\
        & II.6.15a    & 26.14 & 48.11 & 85.30 & 6.62 & 10.61 & 87.42 & \textbf{5.40} & 18.05\\
        & II.11.27    & \textbf{0.83} & \textbf{0.83} & 26.93 & 1.80 & 1.99 & 21.54 & 1.79 & 1.78\\
        & II.11.28    & 4.06 & 30.05 & 24.77 & \textbf{1.66} & 1.97 & 4.58 & 1.99 & \textbf{1.66}\\
        & II.35.21    & 11.78 & 15.08 & 19.09 & 11.72 & 11.89 & 12.30 & \textbf{11.04} & 11.81\\
        & III.9.52    & \textbf{40.90} & 60.28 & 78.22 & 77.72 & 75.60 & 73.26 & 73.01 & 75.81\\
        & III.10.19   & 6.67 & \textbf{3.14} & 10.45 & 5.97 & 14.04 & 10.45 & 6.18 & 7.14\\
        & Pagie   & 90.07 & 63.33 & 1.06e5 & 55.66 & 74.72 & 55.84 & \textbf{54.21} & 55.12\\
        & Kotanchek      & 174.67 & 234.35 & 457 & 252.06 & 387.16 & \textbf{40.06} & 387.16 & 387.16\\
        & UnwrappedBall       & 539.40 & \textbf{153.52} & 287.91 & 864.29 & 934.47 & 1016.75 & 854.92 & 867.21\\
    \end{tabular}}
\end{table*}

The number of rows where the
algorithms with shape constraints are better/equal/worse than
without shape constraints are: $(0, 4, 12)$ without noise, $(7, 0, 9)$
with 10\% noise, $(5, 1, 10)$ with 30\% noise, and $(10, 0, 6)$ with
100\% noise. Without noise the results are again worse with shape constraints.
For the noisy problems the disadvantage is smaller, but
the results do not show that shape constraints improve the results as
strongly as observed for the in-domain problem instances.

\begin{figure}
    \begin{subfigure}[b]{0.5\textwidth}
        \includegraphics[width=\textwidth]{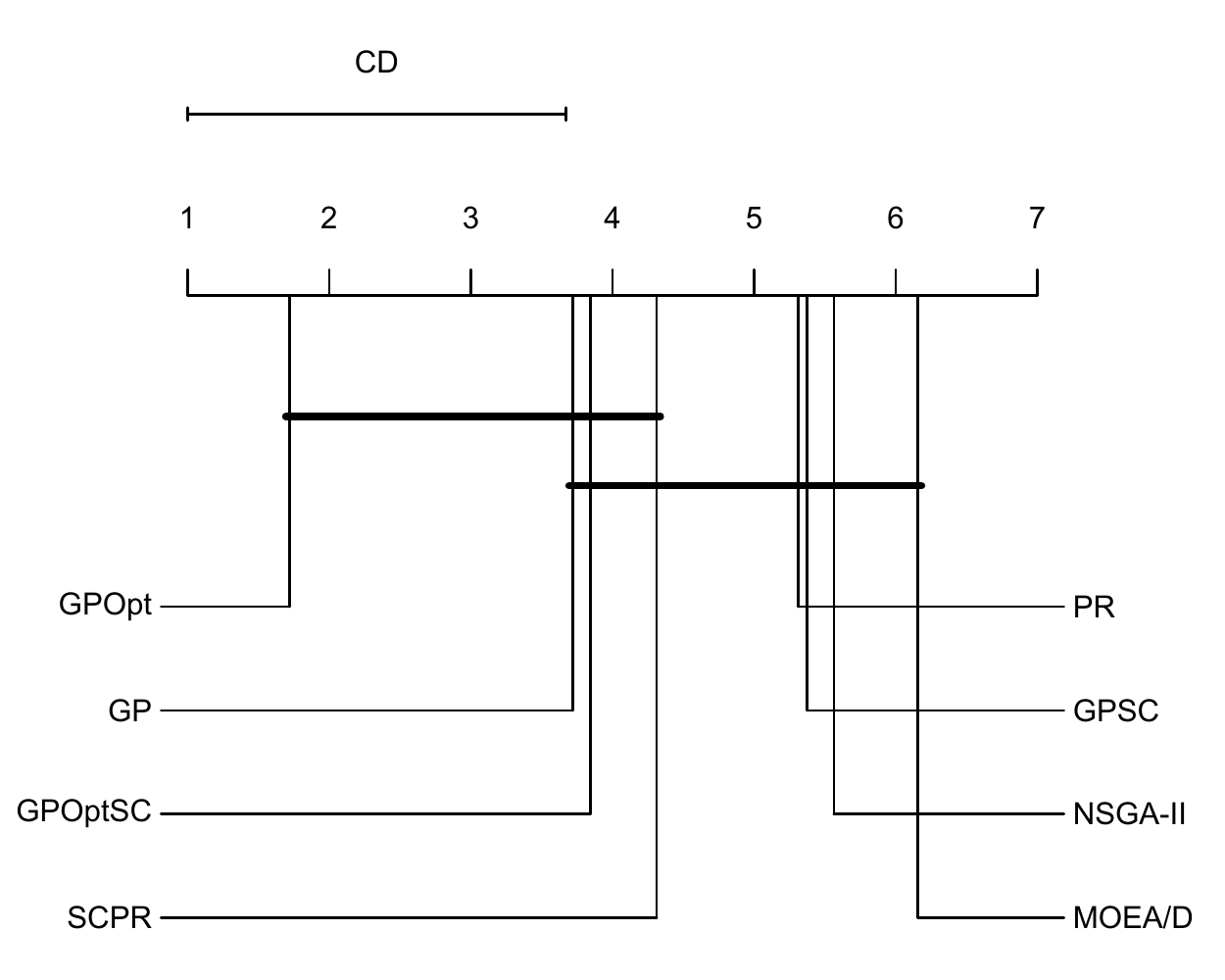} 
        \caption{CD plot on noise level 0\%}
        \label{fig:critical_difference_extrapolation_noise_0} 
    \end{subfigure}
    \begin{subfigure}[b]{0.5\textwidth}
        \includegraphics[width=\textwidth]{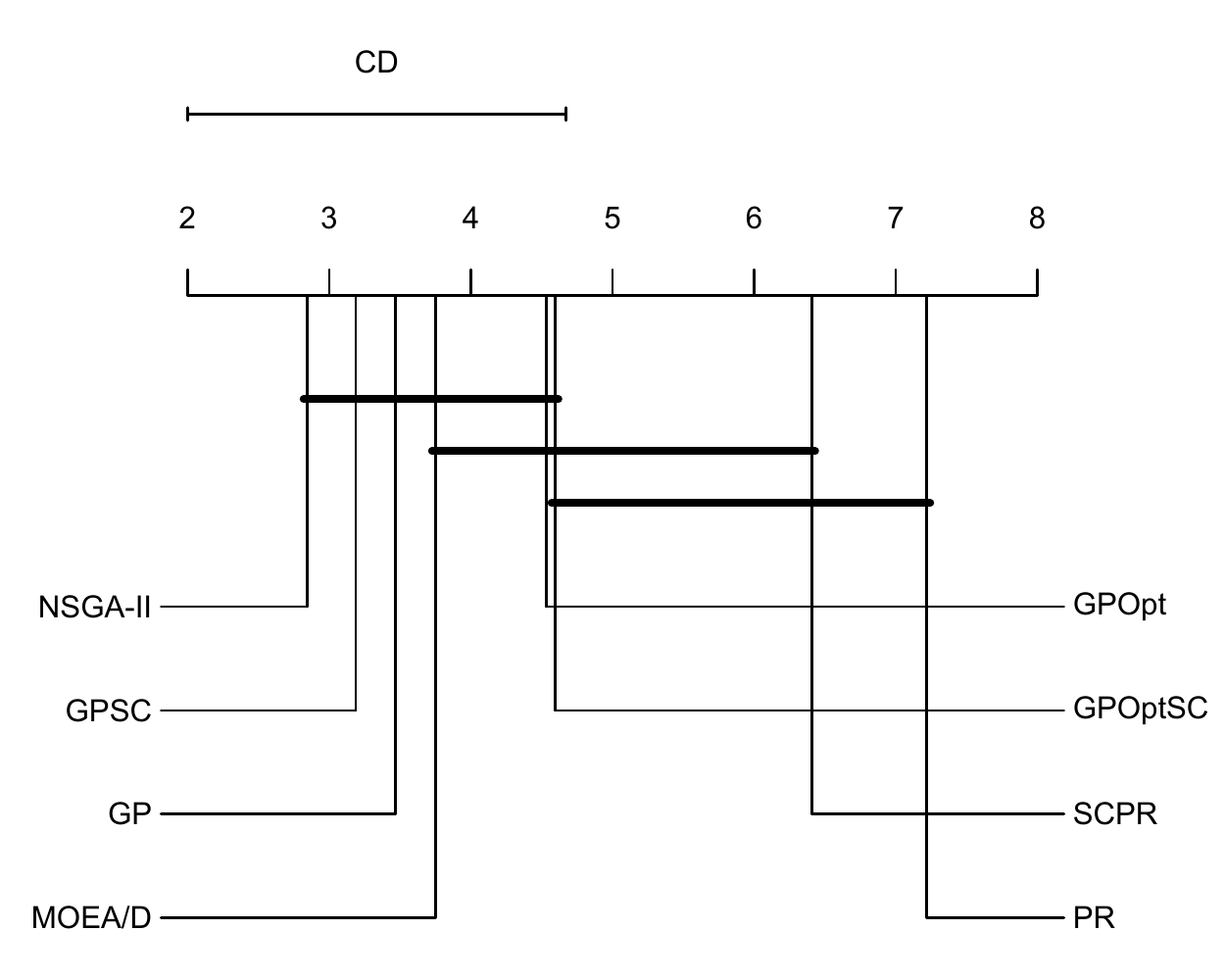}
        \caption{CD plot on noise level 100\%}
        \label{fig:critical_difference_extrapolation_noise_1} 
    \end{subfigure}
    \caption{Critical difference plots for out-of-sample (extrapolation) performance}
    \label{fig:cd_plots_extrapolation}
\end{figure}

Figure \ref{fig:critical_difference_extrapolation_noise_0} shows that GPOpt has
the best rank overall and has significantly better ranking than PR, GPSC,
NSGA-II and MOEA/D. The CD plot for 100\% noise in Figure
\ref{fig:critical_difference_extrapolation_noise_1} shows that again NSGA-II has the
overall best rank and is significantly better than PR and SCPR. 
The CD plots for in-sample and out-of-sample performance are similar.

\section{Discussion}
\label{sec:discussion}


NSGA-II produced on average the best results for the
highest noise, but we found no statistically significant difference
between multi-objective algorithms and the best single-objective
algorithm. However, multi-objective algorithms offer the possibility
to handle each constraint violation separately by using different
solutions from the resulting Pareto front. For the comparison with the
single-objective results, we took the solution with best NMSE from the
Pareto front. 

We did not find a significant difference between MOEA/D
and NSGA-II. MOEA/D may have an advantage with more constraints but we
have not analyzed this in detail so far. SPEA2 and NSGA-III are potentially
also interesting alternatives that could be tried to improve SCSR results.

Inclusion of shape constraints has a small effect on the
runtime. Checking constraint bounds using interval arithmetic is only
slightly more costly than evaluation of the model for one data
point. The calculation of partial derivatives for each solution
candidate is an additional computational burden which depends mainly
on the size of candidate models. We only consider first or
second derivatives, so the additional computational effort is rather small.

Our experiments are focused on synthetic datasets where the underlying
model can be expressed as a short mathematical formula. All problem
instances that we have used have low dimensionality. We do
not know yet how well SCSR works for
high-dimensional problem instances.

We have so far only analyzed a pessimistic approach because we believe
it scales better to high-dimensional problems. However, the problem
instances from the \emph{Feynman Symbolic Regression Database} are all
low-dimensional. For these models an optimistic approach based on
sampling is also computationally feasible. We have not yet compared
the optimistic approach with IA and leave this for future work.

\section{Conclusion}
\label{sec:conclusion}
Our results show that including shape constraints into genetic programming via interval
arithmetic leads to solutions with higher prediction error for
low-noise settings. We have already observed this in our previous
experiments \cite{ourPaper}. The new results including other
problem instances, and multi-objective algorithms again support this conclusion.
With shape constraints we found better models for the higher noise levels,
but we found no statistically significant difference to single-objective genetic
programming without shape constraints.

Multi-objective algorithms for shape-constrained symbolic regression
which do not reject infeasible solutions immediately produce similar
results as single-objective algorithms. On average we found the best
results with NSGA-II but the difference to the other shape-constrained
algorithms was small. We found no significant difference between overall
rankings of NSGA-II and the best algorithm without constraints.

The results for in-domain predictions and out-of-domain predictions
are similar.  We did not find convincing evidence for our hypothesis
that shape constraints are more helpful for extrapolation
(out-of-domain predictions) in general. However, we could show
for a single problem instance that shape constraints produced much better
out-of-domain predictions. The effect is very problem specific.

For the selected problem instances the evolutionary algorithms for
shape-constrained symbolic regression produced better models on
average than shape-constrained polynomial regression which is a
deterministic approach with a strong mathematical background.
However, this requires a computationally costly
grid-search to optimize hyper-parameters for each problem instance.

Still an open topic are the overly wide bounds produced by interval
arithmetic. Alternative methods which produce tighter bounds could
improve evolutionary dynamics and better prediction results. Another interesting idea for future work is to limit the functional
complexity of regression models via shape constraints to limit the
maximum slope or curvature of models.



\section*{Acknowledgment}
The authors gratefully acknowledge support by the Christian Doppler
Research Association and the Federal Ministry of Digital and Economic
Affairs within the Josef Ressel Centre for Symbolic Regression.
This project was partly funded by Funda\c{c}\~{a}o de Amparo \`{a} Pesquisa do Estado de S\~{a}o Paulo (FAPESP), grant number 2018/14173-8.

\bibliographystyle{elsarticle-harv}

\end{document}